\documentclass[journal,draftcls,onecolumn,12pt,twoside]{IEEEtran}
\usepackage{caption}
\usepackage{subcaption}
\usepackage{lipsum}
\usepackage{url}
\usepackage{verbatim}
\usepackage{todonotes}
\usepackage{color}

\usepackage{graphics}
\usepackage{amsmath,amsfonts, amssymb}
\usepackage{mathtools}
\usepackage{epstopdf}
\usepackage{amsthm}
\usepackage{bbm}
\usepackage{algorithm}
\usepackage[noend]{algpseudocode}
\usepackage{dsfont}
\usepackage{cite}
\usepackage{multicol}
\usepackage{multirow}
\usepackage{array}
\usepackage{booktabs}
\usepackage[normalem]{ulem}
\useunder{\uline}{\ul}{}

\newcolumntype{P}[1]{>{\centering\arraybackslash}p{#1}}
\newcolumntype{M}[1]{>{\centering\arraybackslash}m{#1}}

\ifCLASSINFOpdf
\else
\fi
\begin{document}
%
\title{FedADC: Accelerated Federated Learning with Drift Control}
\author{Kerem~Ozfatura,
        Emre~Ozfatura,
        Deniz~G{\"u}nd{\"u}z
\thanks{Kerem Ozfatura is with KUIS AI Lab, Department of Computer Engineering, Koç University, Email: aozfatura22@ku.edu.tr}
\thanks{Emre Ozfatura and Deniz G{\"u}nd{\"u}z are with Information Processing and Communications Lab, Department of Electrical and Electronic Engineering,
Imperial College London Email: \{m.ozfatura, d.gunduz\} @imperial.ac.uk.}
\thanks{This work was supported in part by the Marie Sklodowska-Curie Action SCAVENGE (grant agreement no. 675891), the European Research Council (ERC) Starting Grant BEACON (grant agreement no. 677854), an by UK EPSRC (grant no.EP/T023600/1).}}

\maketitle

\begin{abstract}
Recently, federated learning (FL) has become {\em de facto} framework for collaborative learning among edge devices with privacy concern. The core  of the FL strategy is the use of stochastic gradient descent (SGD) in a distributed manner. Large scale implementation of  FL brings new challenges, such as the incorporation of acceleration techniques designed for SGD into the distributed setting, and  mitigation of the drift problem due to  non-homogeneous distribution of  local datasets. These two problems have been separately studied in the literature; whereas, in this paper, we argue that  it is possible to address both problems using a single strategy without any major alteration to the FL framework, or introducing additional computation and communication load. To achieve this goal, we propose FedADC, which is an accelerated FL algorithm with drift control. We empirically illustrate the advantages of FedADC.
\end{abstract}

\begin{IEEEkeywords}
Drift control, federated learning, momentum SGD, self-knowledge distillation, personalization, classifier calibration, data heterogeneity, regularization.
\end{IEEEkeywords}

%
\IEEEpeerreviewmaketitle

\section{Introduction}
The federated learning (FL) framework has been introduced in \cite{FL1} to enable large-scale {\em collaborative learning} in a distributed manner and without sharing local datasets, addressing, to some extend, the privacy concerns of the end-users. Recently, FL framework has become highly popular, and has been implemented in  several practical applications, such as learning keyboard prediction mechanisms on  edge devices \cite{FL.keyboard1,FL.keyboard2}, or  digital healthcare /remote diagnosis  \cite{FL.health1,FL.health2,FL.health3}.\\
\indent However, large scale implementation of the FL framework introduces new challenges. One of the main obstacles in front of the practical implementation of  FL is that in general the distribution of  data among end users is not homogeneous. As a result, it is possible to observe a noticeable generalization gap in practical scenarios compared to the analysis conducted under  independently and identically distributed (iid) data assumption \cite{FL.noniid1,FL.noniid2,FL.noniid3,FL.noniid4}. The overcome the detrimental affects of such non-iid data distribution, various modifications on the FL framework have been introduced in the recent literature \cite{FL.noniid4,scaffold,fedprox}. In \cite{FL.noniid4}, it is shown that by globally sharing only a small portion of the local datasets, it is possible to tolerate the detrimental affects of  non-iid distribution. However, data sharing, even of a small portion, is contradictory with the fundamental privacy-sensitive learning objective of FL, and may not be possible in certain applications. Alternatively, in \cite{scaffold}, the authors suggest utilizing  stochastic variance reduction \cite{SGD.svr} in the federated setting to control the local drifts due to non-iid data distribution. Another recent approach to control the local drift, studied in \cite{fedprox},  is to penalize the deviation of the local model from the global one.\\
\indent Another line of research explores how to employ acceleration methods in the FL framework. Acceleration methods such as momentum \cite{SGD.nesterov,SGD.polyak}, are known to be highly effective for training  deep neural network (DNN) architectures, increasing both the convergence speed and the final test accuracy  \cite{SGD.opt2}. Recently, in \cite{FL.acc1} and \cite{FL.acc2}, it has been shown that it is possible to employ different acceleration techniques at the server side to achieve better generalization error and to speed up training also in the distributed setting.\\
\indent In this paper, inspired from the previous studies \cite{scaffold} and \cite{FL.acc1}, we introduce accelerated FL with drift control  (FedADC),  which uses {\em momentum stochastic gradient descent (SGD) optimizer} at the server for acceleration as in \cite{FL.acc1}; however, the momentum is updated through  local iterations to prevent local drifts similarly to the SCAFFOLD strategy in \cite{scaffold}, where a globally computed gradient estimate is used to reduce the variance of the local gradient estimates. We want to highlight that the proposed FedADC strategy does not require extra computations or additional hyper-parameters; and hence, can be implemented easily. We need to mention that it requires a momentum vector used by all participating users. Hence, either all the users will receive that, even if they do not participate a round, or (probably more likely) we need to send the momentum together with the model, increasing the communication overhead. However, as we later discuss, this overhead can be overlapped with the computation time to mitigate increase on the communication latency. Next, we briefly summarize the distributed SGD strategy and how it is employed in the FL framework.
Our specific contributions can be summarized as follows:
\begin{itemize}
\item We introduce a {\em momentum embedding} strategy to exploit the global momentum at local client updates to prevent model drift and utilize the local clients updates to update global momentum at the PS to accelerate the learning.
\item We introduce a novel self-knowledge distillation framework, which takes into account  the local data distribution statistics , without sharing with the PS, to regulate the local model updates, and enhance the performance of FedADC framework.
\item By performing extensive simulations in wide range of  scenarios with different dataset and neural network models and comparing with the state-of-the-art schemes, we empirically show the superiority of the proposed framework.
\item We also show that using classifier calibration on top the proposed FedADC framework it is possible obtain personalized models without inducing extra computational cost.
\item Finally, we emphasize that proposed end-to-end framework with personalization, unlike many of the existing alternatives, robust to change of local data statistics.
\end{itemize}

\subsection{Preliminaries}
Consider the following collaborative learning problem across $N$ users, each with its own local dataset denoted by $\mathcal{D}_{i}$:
\begin{equation}
\min_{\boldsymbol{\boldsymbol{\theta}}\in\mathbb{R}^{d}} f(\boldsymbol{\theta})= \frac{1}{N}\sum^{N}_{i=1}\underbrace{\mathds{E}_{\zeta \sim \mathcal{D}_{i}}F(\boldsymbol{\theta},\zeta)}_{\mathrel{\mathop:}=f_{i}(\boldsymbol{\theta})},\label{DSO}
\end{equation}
where $F(\cdot)$ is a parameterized loss function,  $\boldsymbol{\theta}$ is the $d$-dimensional parameter model, $\zeta$ denotes a random sample, and finally $f_{i}(\cdot)$ is the local loss function of the $i$-th users.\\
\indent The parallel stochastic gradient descent  (PSGD) framework is designed to solve the optimization problem in (\ref{DSO}) with the help of a parameter server (PS). At the beginning of each iteration $t$, each user pulls the current parameter model $\boldsymbol{\theta}_{t}$ from the PS, and  computes the {\em local gradient estimate}
\begin{equation}
\nabla_{\boldsymbol{\theta}_{t}}f_{i}(\boldsymbol{\theta}_{\tau},\zeta_{i,t}),
\end{equation}
where $\zeta_{i,t}$ is the  data sampled  by the $i$-th worker from its local dataset at iteration $t$. Then, each worker pushes its local gradient estimate to the PS, where those values are aggregated to update the parameter model, i.e.,
\begin{equation}
\boldsymbol{\theta}_{t+1}= \boldsymbol{\theta}_{t}-\eta_{t}\frac{1}{N}\sum^{N}_{i=1}\nabla_{\boldsymbol{\theta}_{t}}f_{i}(\boldsymbol{\theta}_{t},\zeta_{i,t}),
\end{equation}
where $\eta_{t}$ is the learning rate. In a broad sense, FL  works similarly to PSGD with the following two modifications to address the communication bottleneck. In FL, instead of communicating  each  local gradient estimate to the PS, the users
carry out $H$ local SGD iterations before sending their updated local models to the PS to obtain a consensus model. Moreover, at each iteration of  FL, $cN$ users are chosen randomly for the model update, where $0\leq c \leq 1$ represent the participation ratio. The resultant FL algorithm, also known as Federated Averaging (FedAvg) is provided in Algorithm \ref{alg:fl}, where $\mathcal{S}_{t}$ denotes the set of chosen users at communication round $t$.
\begin{algorithm}[t!]
\caption{Federated averaging (FedAvg) }\label{alg:fl}
\begin{algorithmic}[1]
\For{$t=1,2,\ldots$}
    \State{Choose a subset of users randomly $\mathcal{S}_{t}\subseteq [N]$: $ \vert\mathcal{S}_{t}\vert=cN$} 
    \For{$i\in \mathcal{S}_{t}$}
        \State Pull $\boldsymbol{\theta}_{t}$ from PS: $\boldsymbol{\theta}^{0}_{i,t}=\boldsymbol{\theta}_{t}$
        \For{$\tau=1,\ldots,H$}
            \State Compute SGD: $\mathbf{g}^{\tau}_{i,t}=\nabla_{\boldsymbol{\theta}}f_{i}(\boldsymbol{\theta}^{\tau-1}_{i,t},\zeta_{i,\tau})$
            \State Update model: $\boldsymbol{\theta}^{\tau}_{i,t}=\boldsymbol{\theta}^{\tau-1}_{i,t}-\eta_{t}g^{\tau}_{i,t}$
       \EndFor
       \State Push $\boldsymbol{\theta}^{H}_{i,t}$
    \EndFor
    \State{\textbf{Federated Averaging}:} $\boldsymbol{\theta}_{t+1}=\frac{1}{ \vert \mathcal{S}_{t} \vert}\sum_{i\in\mathcal{S}_{t}} \boldsymbol{\theta}^{H}_{i,t}$
\EndFor
\end{algorithmic}
\end{algorithm}
\subsection{Background and Motivation}
Recently, novel accelerated FL methods have been introduced in \cite{FL.acc1,FL.acc2,FL.acc3} building on the inner/outer loop architecture \cite{lookahead}, where the inner loop involves the local updates and the outer loop is  responsible for the global update. These methods achieve better generalization error compared to the conventional FedAvg strategy. However, robustness of these solutions against the heterogeneity of data in practical FL scenarios is not  discussed in the literature; although it has been demonstrated that employing an additional server-side optimizer can help to alleviate the impact of non-iid data distribution \cite{FL.noniid1}.\\
\indent The impact of  non-iid data distribution on FL has been highlighted recently by several studies \cite{FL.noniid1,FL.noniid2,FL.noniid3,FL.noniid4}. The main challenge in FL, due to non-iid data distribution is {\em local drifts}, which refer to the deviations in the local models from the previous global model due to  iterations based on the local dataset. These local drifts become more prominent over iterations, and lead to a higher generalization gap \cite{FL.noniid1}. Although accelerated FL methods are not particularly designed to mitigate the local drift problem, we argue that, with a slight alteration, server side acceleration methods can be enhanced to be robust against non-iid data distribution without an additional mechanism for drift control. Next, we explain our proposed strategy in detail.

\section{Accelerated Federated Learning with Drift Control (FedADC)}
 The core idea behind our approach is to embed the global momentum update procedure in \cite{FL.acc1} into the local iterations; so that, in a broad sense, the local updates can be considered as a two-player game between the users and the PS, where each decides on the direction of the update alternatively. This way one can enjoy the acceleration offered by the SlOWMO strategy while the local drifts are also controlled.\\
 \indent In the SlOWMO strategy, presented in Algorithm  \ref{alg:SlowMo}, the inner loop, which corresponds to the local updates at the users, is identical to FedAvg in Algorithm \ref{alg:fl}. The key variation lies in the outer loop: unlike in FedAvg, where the local models are averaged to obtain the global model, the PS treats the changes in the local models as {\em pseudo gradients} and utilizes them to compute a global momentum, denoted by $\mathbf{m}_{t}$, which is then used to update the global model.\\
\indent  Formally speaking, at the beginning of  each communication round $t$, each selected user pulls the latest global model, $\boldsymbol{\theta}_{t}$, from the PS, performs $H$ local updates, and sends the  accumulated model update $\boldsymbol{\Delta}_{n,t}$ to the PS (as illustrated in line 10 of Algorithm \ref{alg:SlowMo}). The PS utilizes the average of the model updates as the global pseudo gradient, and updates the momentum  of the outer loop accordingly (illustrated in line 14 of Algorithm \ref{alg:SlowMo}). Finally, the PS \footnote{SLOWMO framework is actually designed for the DSGD setup, so that the model updates are performed locally by using all-reduce communication, here we present its adapted to version to FL setup.} updates the global model using the momentum just obtained (line 16 of Algorithm \ref{alg:SlowMo}).\\
\begin{algorithm}[t!]
\caption{SLOWMO}\label{alg:SlowMo}
\begin{algorithmic}[1]
    \For{$t=1,\ldots,T$}
    \State \underline{\textbf{Local iteration:}}
     \For{$i\in\mathcal{S}_{t}$} in parallel
      \State $\boldsymbol{\theta}^{0}_{i,t}=\boldsymbol{\theta}_{t}$
        \For{$\tau=1,\ldots,H$} local update:
            \State Compute SGD: $\mathbf{g}^{\tau}_{i,t}=\nabla_{\boldsymbol{\theta}}f_{i}(\boldsymbol{\theta}^{\tau-1}_{i,t},\zeta^{\tau}_{i,t})$
            \State Update model: $\boldsymbol{\theta}^{\tau}_{i,t}=\boldsymbol{\theta}^{\tau-1}_{i,t}-\eta_{t}g^{\tau}_{i,t}$
       \EndFor
    \EndFor
\State\underline{\textbf{Communication phase:}}
\For{$i\in\mathcal{S}_{t}$}
\State{Send $\boldsymbol{\Delta}_{i,t}=\boldsymbol{\theta}_{t}-\boldsymbol{\theta}^{H}_{i,t}$ to PS}
\EndFor
\State \underline{\textbf{Compute pseudo gradient:}}
\State $\bar{\mathbf{g}}_{t} = \frac{1}{\vert\mathcal{S}_{t}\vert}\frac{1}{\eta_{t}}\sum_{n\in\mathcal{S}_{t}} \boldsymbol{\Delta_{n,t}}$
\State \underline{\textbf{Compute pseudo momentum:}}
\State $\mathbf{m}_{t+1}=\beta\mathbf{m}_{t} + \bar{\mathbf{g}}_{t}$
\State\underline{\textbf{Model update:}}
\State $\boldsymbol{\theta}_{t+1}=\boldsymbol{\theta}_{t}-\alpha\eta_{t}\mathbf{m}_{t+1}$
\EndFor
\end{algorithmic}
\end{algorithm}
We want to remark to that in the SLOWMO framework a second momentum can also be used for the local updates;  however, in the federated setting, as the skewness of the data distribution increases, this local momentum makes the impact of the non-iid distribution even more severe.\\ 
\indent Next, we introduce FedADC, which benefits from the SLOWMO framework with the additional robustness against non-iid data distribution. The key design trick we use here is to embed the momentum update part (illustrated in line 14) into the local iteration; that is, instead of adding the momentum term $\mathbf{m}_{t}$ to the pseudo gradient at the end of the communication round, it is first normalized with respect to the number of local iterations, $\bar{\mathbf{m}}_{t}=\mathbf{m}_{t}/H$, then added to the pseudo gradient gradually through local iterations as illustrated in Algorithm \ref{fedadc}. In Algorithm \ref{fedadc}, local model updates are treated as a two-player game between the user and the PS whose actions correspond to choosing the direction of the model update. The user decides on its action based on the local gradient estimate as (lines 8 and 10) while the PS decides based on the previous global pseudo momentum $\mathbf{m}_{t}$, specifically  the normalized pseudo momentum $\bar{\mathbf{m}}_{t}$. By virtue of this mechanism, each worker searches for the {\em minima} based on its local loss function, while at the same time the PS pulls the local model towards the previous consensus direction to confine the local drift.\\
\indent We consider two variations for the local updates in Algorithm \ref{fedadc} based on whether the actions are taken simultaneously or consecutively, illustrated with {\color{blue}blue} and {\color{red}red} respectively. One can observe that the variation illustrated with red resembles the Nesterov momentum strategy whereas the one with blue resembles heavy ball momentum \cite{SGD.opt2}.\\
\begin{algorithm}[t!]
\caption{Accelerated FL with drift control (FedADC)}\label{fedadc}
\begin{algorithmic}[1]
    \For{$t=1,\ldots,T$}
    \State \underline{\textbf{Local iteration:}}
     \For{$i\in\mathcal{S}_{t}$} in parallel
      \State $\boldsymbol{\theta}^{0}_{i,t}=\boldsymbol{\theta}_{t}$
      \State $\bar{\mathbf{m}}_{t}=\beta_{local}\mathbf{m}_{t}/H$
       \For{$\tau=1,\ldots,H$} local update:
            \State {\color{red} $\boldsymbol{\theta}^{\tau-1/2}_{i,t}=\boldsymbol{\theta}^{\tau-1}_{i,t}-\eta_{t}\bar{\mathbf{m}}_{t}$}
            \State {\color{red}$\mathbf{g}^{\tau}_{i,t}=\nabla_{\boldsymbol{\theta}}f_{i}(\boldsymbol{\theta}^{\tau-1/2}_{i,t},\zeta^{\tau}_{i,t})$}
            \State {\color{red}$\boldsymbol{\theta}^{\tau}_{i,t}=\boldsymbol{\theta}^{\tau-1/2}_{i,t}-\eta_{t}\mathbf{g}^{\tau}_{i,t}$}
              \State {\color{blue} $\mathbf{g}^{\tau}_{i,t}=\nabla_{\boldsymbol{\theta}}f_{i}(\boldsymbol{\theta}^{\tau-1}_{i,t},\zeta^{\tau}_{i,t})$}
            \State {\color{blue} $\boldsymbol{\theta}^{\tau}_{i,t}=\boldsymbol{\theta}^{\tau-1}_{i,t}-\eta_{t}(\mathbf{g}^{\tau}_{i,t}+\bar{\mathbf{m}}_{t})$}
       \EndFor
    \EndFor
\State\underline{\textbf{Communication phase:}}
\For{$i\in\mathcal{S}_{t}$}
\State{Send $\boldsymbol{\Delta}_{i,t}=\boldsymbol{\theta}_{t}-\boldsymbol{\theta}^{H}_{i,t}$ to PS}
\EndFor
\State \underline{\textbf{Compute Pseudo momentum:}}
\State $\bar{\boldsymbol{\Delta}}_{t}=\frac{1}{\vert\mathcal{S}_{t}\vert}\frac{1}{\eta_{t}}\sum_{i\in\mathcal{S}_{t}} \boldsymbol{\Delta}_{i,t}$
\State $\mathbf{m}_{t+1}= \bar{\boldsymbol{\Delta}}_{t} +(\beta_{global}-\beta_{local})\mathbf{m}_{t}$
\State\underline{\textbf{Model update:}}
\State $\boldsymbol{\theta}_{t+1}=\boldsymbol{\theta}_{t}-\alpha\eta_{t}\mathbf{m}_{t+1}$
\EndFor
\end{algorithmic}
\end{algorithm}
We want to stress that the overall update direction over $H$ iterations, $\boldsymbol{\Delta}_{i,t}=\boldsymbol{\theta}_{t}-\boldsymbol{\theta}^{H}_{i,t}$, can be written in the following form:
\begin{equation}
\boldsymbol{\Delta}_{i,t}=\eta_{t}\left(\sum^{H-1}_{\tau=0}\mathbf{g}^{\tau}_{i,t} +\beta_{local}\mathbf{m}_{t}\right),
\end{equation}
and the average of  all the  local updates is given by
\begin{equation}\label{avg_update}
\frac{1}{\vert\mathcal{S}_{t}\vert}\sum_{i\in\mathcal{S}_{t}}\boldsymbol{\Delta}_{i,t}=\eta_{t}\left(\frac{1}{\vert\mathcal{S}_{t}\vert}\sum_{i\in\mathcal{S}_{t}}\sum^{H-1}_{\tau=0}\mathbf{g}^{\tau}_{i,t} +\beta_{local}\mathbf{m}_{t}\right).
\end{equation}
If the both sides of Equation (\ref{avg_update}) are divided by $\eta_{t}$, the right hand side of the equation is in the form  $\beta_{local}\mathbf{m}_{t} + \bar{\mathbf{g}}_{t}$. Finally, one can observe that with a small correction, by adding $(\beta_{global}-\beta_{local})\mathbf{m}_{t}$, the definition in (\ref{avg_update}) becomes identical to the pseudo momentum of the SLOWMO framework. Therefore, although the mechanism for the local updates is modified, the structure of the outer loop closely resembles the SLOWMO strategy.\\
\indent We note that, in Algorithm \ref{fedadc}, we use one additional discount parameter for the momentum term to specify how it is embedded to local iterations. By playing with $\beta_{local}$ parameter it is possible to seek different strategies. However, to keep the number of hyper-parameters same with the FedAvg scheme, we assume $\beta_{local}=\beta_{global}=\beta$.\\
\indent Overall, the proposed FedADC strategy uses  $\mathbf{m}_{t}$ for drift control in the inner loop and for acceleration in the outer loop. At this point we further argue that, although the use of $\mathbf{m}_{t}$ in the inner loop is not similar to the momentum optimizer framework, it may still serve the common purpose of helping to escape from saddle points. In \cite{SGD.opt1}, it has been argued, backed by some theoretical analysis under certain assumptions (which can be validated through experiments) that the efficiency of the momentum approach is due to perturbation of the model parameters towards the escape direction from a saddle point. Hence, we argue that the fixed perturbation $\bar{\mathbf{m}}_{t}$, which does not depend on the local gradient estimate, may also help to escape from a saddle point. Around saddle point, the local gradient estimates start to vanish; however, since the term $\bar{\mathbf{m}}_{t}$ does not depend on the local gradient estimates, vanishing behavior will not be observed for $\bar{\mathbf{m}}_{t}$ term; and thus, the perturbation due to $\bar{\mathbf{m}}_{t}$ will prevent the local model from  being stuck at a saddle point.

\begin{algorithm}[t!]
\caption{FedADC with double momentum}\label{double_momentum}
\begin{algorithmic}[1]
    \For{$t=1,\ldots,T$}
    \State \underline{\textbf{Local iteration:}}
     \For{$i\in,\mathcal{S}_{t}$} in parallel
      \State $\boldsymbol{\theta}^{0}_{i,t}=\theta_{t}$
      \State $\bar{\mathbf{m}}_{t}=\beta\mathbf{m}_{t}/H$
        \For{$\tau=1,\ldots,H$} local update:
        \State $\mathbf{g}^{\tau}_{i,t}=\nabla_{\boldsymbol{\theta}}f_{i}(\boldsymbol{\theta}^{\tau-1}_{i,t},\zeta^{\tau}_{i,t})$
            \State \textbf{Update local momentum}:
            \If{$\tau=1$}
            \State $\mathbf{m}^{\tau}_{i,t}=\mathbf{g}^{\tau}_{i,t}$
            \Else
            \State $\mathbf{m}^{\tau}_{i,t}= (\mathbf{m}^{\tau-1}_{i,t}\phi+(1-\phi)\mathbf{g}^{\tau}_{i,t})$
            \EndIf
            \State \textbf{Update local model}:
            \State $\boldsymbol{\theta}^{\tau}_{i,t}=\boldsymbol{\theta}^{\tau-1}_{i,t}-\eta_{t}(\bar{\mathbf{m}}_{t}+\mathbf{m}^{\tau}_{i,t})$
       \EndFor
    \EndFor
\State\underline{\textbf{Communication phase:}}
\For{$i\in\mathcal{S}_{t}$}
\State{Send $\Delta_{i,t}=\theta_{t}-\boldsymbol{\theta}^{H}_{i,t}$ to PS}
\EndFor
\State \underline{\textbf{Compute average diffrence:}}
\State $\bar{\boldsymbol{\Delta}}_{t} = \frac{1}{\vert\mathcal{S}_{t}\vert}\frac{1}{\eta_{t}}\sum_{i\in\mathcal{S}_{t}} \Delta_{i,t}$
\State \underline{\textbf{Compute Pseudo momentum:}}
\State $\mathbf{m}_{t+1}= \bar{\boldsymbol{\Delta}}_{t}$
\State\underline{\textbf{Model update:}}
\State $\boldsymbol{\theta}_{t+1}=\boldsymbol{\theta}_{t}-\alpha\eta_{t}\mathbf{m}_{t+1}$
\EndFor
\end{algorithmic}
\end{algorithm}

\indent We remark that momentum that employed in the FedADC  is directly embedded from the global momentum while local client does not employ a personal momentum. To this end, we introduce a complete version of FedADC method called FedADC with double momentum, where we also incorporate the local momentum for the clients at local iteration in Algorithm \ref{double_momentum}. Due to the additional $\phi$ hyperparameter that use for scaling the local momentum \textrm{$\boldsymbol{m}$}, in our simulations we consider the FedADC method without the double momentum to reduce the amount of hyperparameters.    
\subsection{Communication load}
One of the main design goals behind the FedADC strategy is to keep the number of hyper-parameters and the number of exchanged model parameters as low as possible while accelerating the training and achieving a certain robustness against non-iid data distribution and acceleration. From the uplink perspective, proposed strategy does not impose any additional communication load compared to conventional FedAvg strategy; however, on the downlink, the model difference $\bar{\boldsymbol{\Delta}}_{t}$ should be broadcasted to all the users  in the system to ensure that each user  can track the global momentum $\mathbf{m}_{t}$.\\
\indent Alternatively, at each global update phase, selected users in $\mathcal{S}_{t}$ can pull both the global momentum $\mathbf{m}_{t+1}$ and the global model $\boldsymbol{\theta}_{t}$ form the PS, which doubles the communication load in the downlink direction. However, in certain settings the additional communication load can be hidden by overlapping it with the computation time, which would  prevent an increase in the overall communication latency. To clarify, at time slot $t$, in parallel to the computation process of users in $\mathcal{S}_{t}$, the PS can decide  the next set of users, $\mathcal{S}_{t+1}$, and send them the current momentum $\mathbf{m}_{t}$ and the model $\boldsymbol{\theta}_{t}$; hence, at the beginning of  iteration $t+1$, the users in $\mathcal{S}_{t+1}$ only need to pull the average model difference $\bar{\boldsymbol{\Delta}}_{t}$ at the beginning of  iteration $t+1$. Hence, although the additional communication load at the downlink side cannot be prevented, the corresponding latency can be reduced by overlapping it with the computation time. 
\section{FedADC with Self Knowledge Distillation}
\label{conf-KD}
The recent works \cite{selfkd1,selfkd2,selfkd3,selfkd4}, has shown that {\em self knowledge distillation} works as a regularizer during the training which  mitigates unintended memorization and hence improves the generalization. In a broad sense, the main idea behind the self knowledge distillation is to define a {\em target probability} for the training data, which dynamically adjusted according to the current prediction, and utilize it for training instead of true labels. Consequently, use of hand tailored target probabilities makes the learning process smoother and improves the generalization.\\
\indent The same notion can be also employed in FL setup to mitigate local drift. In this case the global model at the PS can be used to define target probabilities to prevent over-fitting to local data. This strategy  has been recently investigated in  \cite{dist1} and \cite{dist2}. Different from \cite{dist1}, in \cite{dist2} the authors consider the knowledge distillation for only non-true classes so that knowledge distillation do not slows down the local learning process. Prior to FL, similar approach has been already considered in the literature of adversarial learning to prevent robust over-fitting \cite{distadv} and has shown noticeable improvement.\\

\begin{equation}\label{eq:KL}
\left\{\begin{matrix}
p_{i}(\mathbf{x}\vert\boldsymbol{\theta}) = \frac{\textup{exp}(y_{i}/\tau)}{\sum_{j}\textup{exp}(y_{j}/\tau)}
\\ 
\hat{p}_{i}(\mathbf{x}\vert\boldsymbol{\theta_{T}}) = \frac{\textup{exp}(\hat{y}_{i}/\tau)}{\sum_{j}\textup{exp}(\hat{y}_{j}/\tau)}
\end{matrix}\right.
\quad, \qquad
\mathcal{L}_{KL}(\mathbf{p},\hat{\mathbf{p}}) = -\sum_{i}\hat{p}_{i}\;{\textup{log}}\left ( \frac{p_{i}}{\hat{p}_{i}} \right ).
\end{equation}
We give data $\mathbf{x}$ to model $\boldsymbol{\theta}$ and write $p_{i}(\mathbf{x}\vert\boldsymbol{\theta})$ which stands for prediction probability on on class $i$ as a softmax function of logits $y$. KL divergence can utilize temperature coefficient $\tau$ to divide logit values for tuning the prediction probabilities. $\mathcal{L}_{KL}$ is the KL-Divergence loss between student and teacher prediction.\\
\indent The knowledge distillation strategy for FL utilizes a loss function $\mathcal{L}$ that combines the cross-entropy (CE) loss with a regularizing loss function. The common choice for regularizing loss that utilizes the notion of knowledge distillation is KL divergence which penalize the difference between two probability distributions.
The structure of the overall loss function is illustrated in equation (\ref{lossfunc}),

\begin{equation}\label{lossfunc}
\mathcal{L}=(1-\lambda)\mathcal{L}_{CE}(f{(\mathbf{x\vert}\boldsymbol{\theta})},y) + \lambda\mathcal{L}_{KL}(p({\mathbf{x}\vert\boldsymbol{\theta})},\hat{\mathbf{p}};\tau)
\end{equation}
where $(\mathbf{x},y)$ is  the pair of data and the corresponding label, $f(\mathbf{x}\vert{\boldsymbol{\theta}})$ is the logit values corresponding to $\mathbf{x}$ based on the local model $\boldsymbol{\theta}$, $p(\mathbf{x}\vert{\boldsymbol{\theta}})$ is the probability vector corresponding to $\mathbf{x}$ based on the local model $\boldsymbol{\theta}$, 
$\hat{\mathbf{p}}$ is the target probability vector, and $\lambda\in[0,1]$ is the weight to combine the two loss functions.\\
\indent The main challenge in self knowledge distillation is to define the target probability $\hat{\mathbf{p}}$. In \cite{dist1}, the authors consider global model at the PS, $p({\boldsymbol{\theta_{ps}}}\vert\mathbf{x})$, to obtain the target probability in order to prevent forgetting global model through the local iterations. However, the main limitation of the proposed schemes in \cite{dist1,dist2} is that the knowledge distillation framework do not utilize the distribution of the local dataset. To this end, we propose a new knowledge distillation strategy, which takes into account the local data distribution. We referred this scheme as {\em self confidence knowledge distillation}.\\ 
\indent Let $\gamma_{i,k}$ be the proportion of the data belong to $i$th class at client $k$. Let $\gamma^{max}_{k}$ denote the largest portion. Then we define $\rho_{i,k}=\frac{\gamma_{i,k}}{\gamma^{max}_{k}}$ as the confidence parameter. The notion behind the use of confidence parameter is to seek a proper way of combining local and global prediction based on the local data distribution, and hence overcome the unintended forgetting of global model. Accordingly, we define  the target probability for knowledge distillation in the following way
\begin{equation}
\hat{p}_{i,k}=(1-\rho_{i,k})\tilde{p}^{(i)}_{\boldsymbol{\theta}}
\end{equation}
for non true class $i\neq y$ and
\begin{equation}
\hat{p}_{j,k}=1-\sum_{i\neq j}(1-\rho_{i,k})\tilde{p}^{(i)}_{\boldsymbol{\theta}}
\end{equation}
for true class $j=y$ \\
We remark that when data is iid distributed, one can observe that $\rho_{i,k}\approx1$ for all non-true class label $i$ and thus the overall loss will be almost identical to CE loss.\\
\indent We combine the FedADC framework with self-knowledge distillation strategy by simply employing the loss function in (\ref{lossfunc}), and we refer the overall method as FedADC${+}$. For the implementation we follow the implementation \footnote{\url{https://github.com/alinlab/cs-kd}} in \cite{selfkd2}.\\
\indent For the completeness of the related literature, in \cite{ensemble1}, the authors explore the idea  of using distillation mechanism at the PS to successfully ensemble the local models instead of simple averaging, however, it requires additional data. Later, in \cite{ensemble3} the authors address this problem by employing a generator do remedy need for proxy data.
\\

\indent Finally, we want to remark that the  underlying reason to introduce a new distillation to mechanism instead of employing the existing ones is their lack of adaptivity to data heterogeneity. To clarify, the performance of the existing distillation mechanisms to prevent unintended forgetting during local learning  requires hyper parameter tuning based on data heterogeneity, however, since the proposed distillation mechanism already takes into account data distribution the hyper-parameter will have less impact on the performance which is desired for the practical implementations. We also want to emphasize that the proposed distillation strategy aim to address the unintended forgetting due to class in-balance among the clients, however, the proposed mechanism can be extended to further address data-size in-balances among clients.

\section{Numerical results}
\subsection{Implementation details}
 Here we further explain how we implemented the other algorithms that we use for the comparison. For all experiments we grid search learning rate values of [0.1,0.05,0.03] with exception of moon algorithm and weight decay values of [0,1e-5,5e-5,1e-4] and reported the their best result respect to the simulation type.
 \subsubsection{FedRS}
 In fedRS\cite{FedRS} algorithm, during trainig, local client's predictions for a label that is not included clients local training data are scaled with the $\alpha$ parameter. Hence we only include FedRS in sort and partition simulations, since in dirichlet simulations there is no restriction to distribute from any image class. In our simulations we use $\alpha=0.5$ and set local momentum to 0.9.
 
\subsubsection{MOON}
MOON\cite{Moon} algorithm requires an additional projection layer in the neural network to employ contrastive learning to direct the local training of the clients. Projected local previous model features are seen as negative while projected global model features are seen as positive samples. For experiments that train on CNN, we remove last two linear layers at the end of the network and swap it with average pooling to reduce the image size following with a projection layer of 2 layer MLP with output dimension of 256. For ResNet architectures, we inject projection layer that is identical to authors of \cite{Moon}. We set temperature to $T=0.5$, learning rate to $0.01$ and weight decay to $1e-5$ and do grid search to find the best $\mu$ values from $[1,5]$. 
\subsubsection{FedGKD}
FedGKD\cite{dist2} algorithm employs knowledge distillation by considering global model as a teacher model during the training. We set $T=0.5$ for temperature and the coefficient to control the scale for knowledge distillation loss $\lambda=0.1$, set the local momentum to $0.9$ as the authors stated in their paper.
\subsubsection{FedLS-NTD}
FedLS-NTD\cite{dist1} algorithm employs knowledge distillation between logit values of the local client and parameter server for the non-true classes. For temperature and distillation loss coefficients, we use $T=1$, $\beta=0.3$  respectively and set the local momentum to 0.9 as the authors stated in their paper.
\subsubsection{FedProx}
Fedprox\cite{fedprox} algorithm considers regularization of proximal term which is scaled with $\mu$ coefficient before adding to the local gradient values. We grid search optimal the $\mu$ from the values [0.1,0.01,0.001] and set local momentum to 0.
\subsubsection{FedDYN}
FedDyn\cite{FedDyn} algorithm considers a dynamic regularization using local client and global model to dynamically compute local gradients, thus ensures the local optima consistency. We grid search the $\alpha$ coefficient values from [0.1, 0.01, 0.001] which is use for scaling the dynamically penalized risk for local gradient values. Weight decay is set to $1e-3$ as the authors stated in their paper

\subsection{Accelerating FL}
\subsubsection{Simulation setup}
\label{setup:1}
For the experiments, we use the CIFAR-10 \cite{cifar10} image classification dataset, which contains 50,000 training and 10,000 test images from 10 classes. To analyze the performance of the given schemes under non-iid data distribution, we consider the {\em sort and partition} approach  to distribute the training dataset among the users in with training images are equally distributed among 100 users. We consider a  neural network architecture with 4 convolutional layers and 4 fully connected layers with no batch normalization applied to the outputs of the layers. Maxpooling is also utilized for the scaling down the image size. We set weight decay to $4\times10^{-4}$.
\subsubsection{Simulation results }
\begin{figure}[h]
    \begin{subfigure}{0.32\textwidth}
        \includegraphics[width=\linewidth]{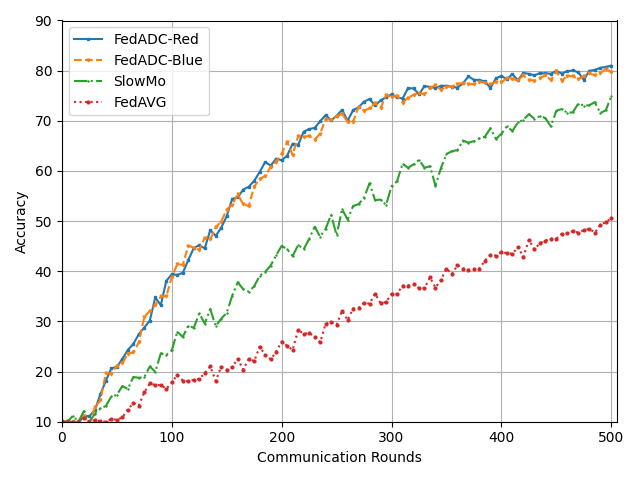}
        \caption{$s=2$}
        \label{s2}
    \end{subfigure}
    \hfill
    \begin{subfigure}{0.32\textwidth}
        \includegraphics[width=\linewidth]{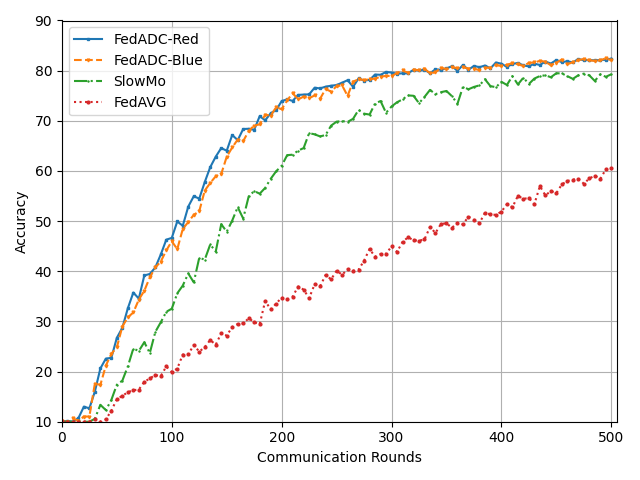}
        \caption{$s=3$}
        \label{s3}
    \end{subfigure}
    \hfill
    \begin{subfigure}{0.32\textwidth}
        \includegraphics[width=\linewidth]{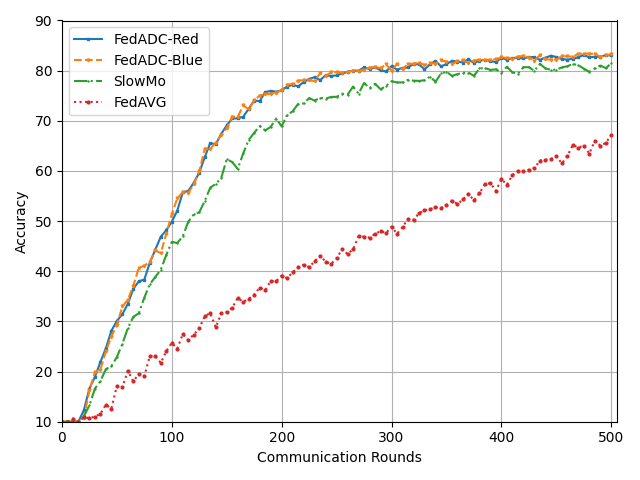}
        \caption{$s=4$}
        \label{s4}
    \end{subfigure}
    \caption{Comparison of the test accuracy of FedADC, FedAvg and  SlOWMO schemes}
    \label{fig:cifar-10-accel}
\end{figure}

 In the sort and partition approach, the training dataset is initially sorted based on the labels, and then they are divided into blocks and distributed among the users randomly based on a parameter $s$, which measures the skewness of the data distribution. To be more precise, $s$ defines the maximum number of different labels within the dataset of the each user, and therefore, the smaller $s$ is, the more skewed the data distribution is. In our numerical experiments, we consider three different scenarios for the skewness of the data distribution with $s=2,3,4$, respectively. In all the experiments we train the given DNN architecture for $500$ communication rounds,  each of which consists of $H=8$ local iterations. We fixed the user participation ratio to $C=0.2$; that is, at each iteration only 20 users participate the training. Further,  we use the batch size of 64.\\
 \indent We consider FedAvg and SlOWMO frameworks as benchmark strategies. To provide a fair comparison we tune all the hyper-parameters including the learning rate $\eta$ and the momentum coefficient $\beta$ for each framework separately. For this we carry out a grid search over the values of $\eta\in\left\{0.01,0.025,0.05,0.1\right\}$ and $\beta\in\left\{0.6,0.7,0.8,0.9\right\}$. We fix  $\alpha=1$ similarly to the \cite{FL.acc1}. For the FedADC we implement both variation illustrated with  {\color{blue}blue} and {\color{red}red} in Algorithm \ref{fedadc}. Finally, we remark that each experiment is repeated 10 times and the average test accuracy results are illustrated.\\
 \indent The convergence results of the FedADC, FedAvg and SlOWMO schemes under non-iid data distribution for $s=2,3,4$ are illustrated in Fig.s \ref{s2}-\ref{s4}, respectively. One can clearly observe that SlOWMO provides a significant improvement over FedAvg, but in all three scenarios, the proposed FedADC scheme outperforms the SlOWMO. The improvement of SlOWMO with respect to FedAvg is consistent with the observations in \cite{FL.noniid1} and \cite{FL.acc1} that the use of {\em server side momentum} can help to mitigate the impact of local drifts, and it  also  accelerates learning. However, by comparing gap between the proposed FedADC framework and SlOWMO for different $s$ values, we also observe that SlOWMO mainly serves for acceleration rather than a drift control mechanism, thus the performance gap between FedADC and SlOWMO widens as the parameter $s$ decreases, i.e., as the data distribution becomes more skewed.\\
 
\begin{figure}[h]
\begin{center}
\includegraphics[scale =0.35]{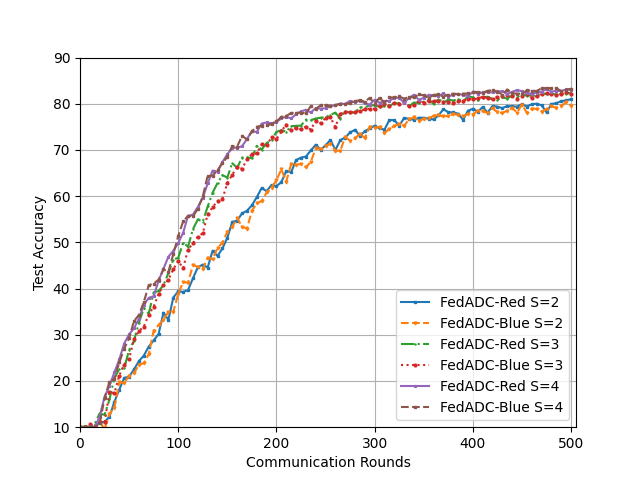}
\caption{Convergenge performance of FedADC  for $s=2,3,4$.}
\label{ADC}
\end{center}
\end{figure}

\indent Finally, in Fig. \ref{ADC}, we compare the test accuracy  of the FedADC scheme for different $s$ parameters to investigate its robustness against non-iid data distribution. We observe that in all the cases FedADC passes \%80 test accuracy within  500 communication rounds. Besides, the simulation results  indicate that, although the convergence speed slows down as $s$ decreases, it seems FedADC still converges to a similar test accuracy level for different $s$ values, which shows the robustness of the FedADC scheme to non-iid data distribution. We also oberve that when $s$ is larger, the two local update mechanisms illustrated with {\color{red} red} and {\color{blue} blue} in Algorithm \ref{fedadc} perform almost identical however as $s$ decreases Nesterov-type model updates, illustrated with {\color{red} red}, slightly performs better as one can observe in Fig. \ref{ADC}.

\subsection{Local drift control and comparison with SOTA Methods}
\subsubsection{Simulation setup}
\label{sec:drift_setup}
To analyze the performance of the given schemes under non-iid data distribution, on CIFAR-10 experiments we use the sort and partition and set the data distribution to $s=2$ class per client with using the same CNN architecture that we use for the acceleration section \ref{setup:1}. For CIFAR-100 experiments, we use the Dirichlet distribution Dir($\alpha$) to sample disjoint non-IID client training data where $\alpha$ denotes the heterogeneity parameter, and the smaller $\alpha$ result in higher data skewness and we consider ResNet-18 \cite{ResNet} neural network architecture and apply group normalization \cite{GroupNorm} after convolutional layers with group size of 32. We set the weight decay to 5e-5 and 5e-4 for Cifar-10 and Cifar-100 respectively. For FedADC+, we set distillation coefficient as $\lambda=0.35$. Further, use batch size of 50 for all simulations. We run all the methods four trials and report the mean and standard derivation.

\subsubsection{Simulation results}
\begin{figure}[h!]
     \centering
     \begin{subfigure}[b]{0.48\textwidth}
         \centering
         \includegraphics[width=\textwidth]{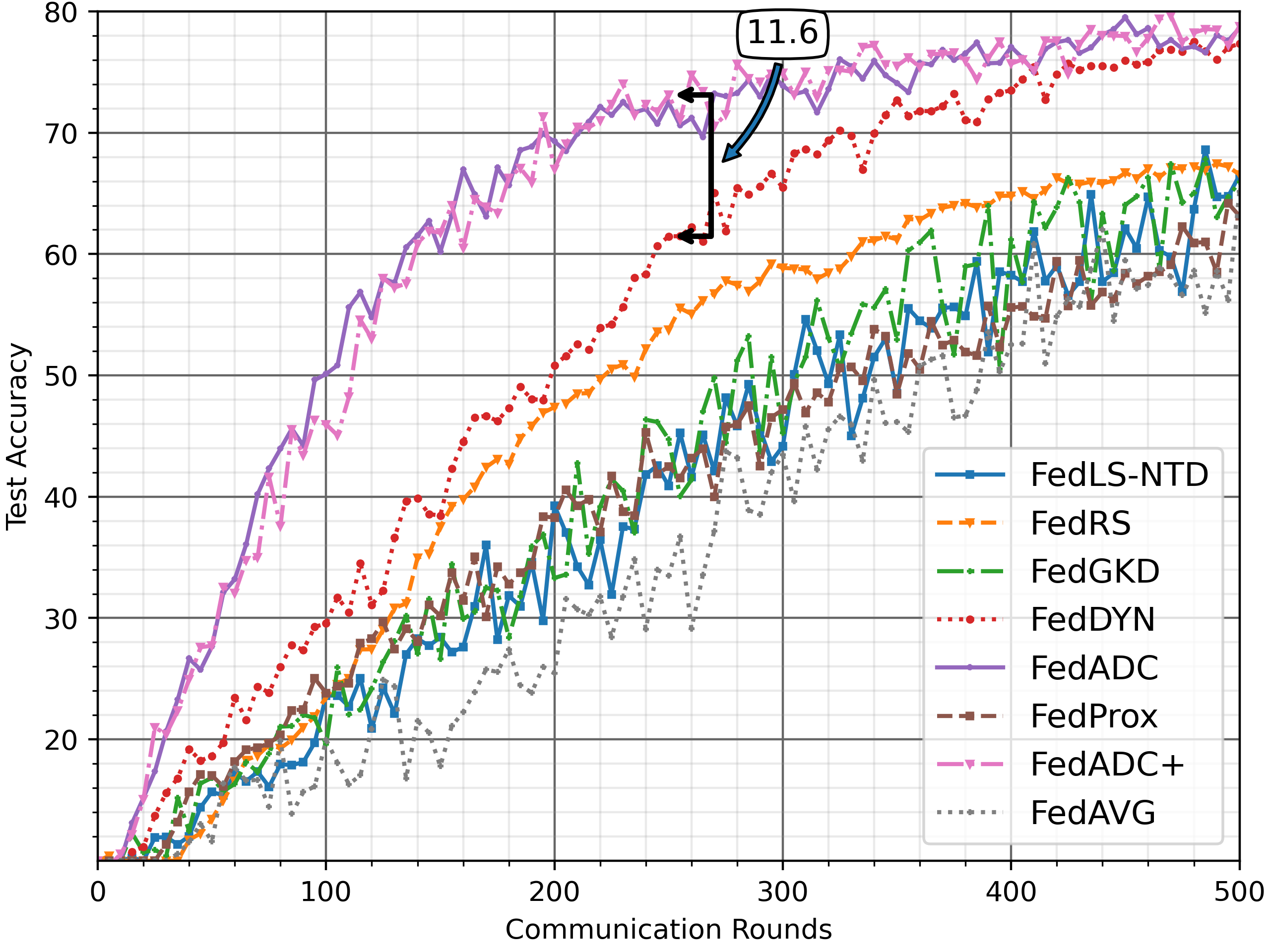}
         \caption{$C=0.1$}
         \label{cif10-01}
     \end{subfigure}
     \hfill
     \begin{subfigure}[b]{0.48\textwidth}
         \centering
         \includegraphics[width=\textwidth]{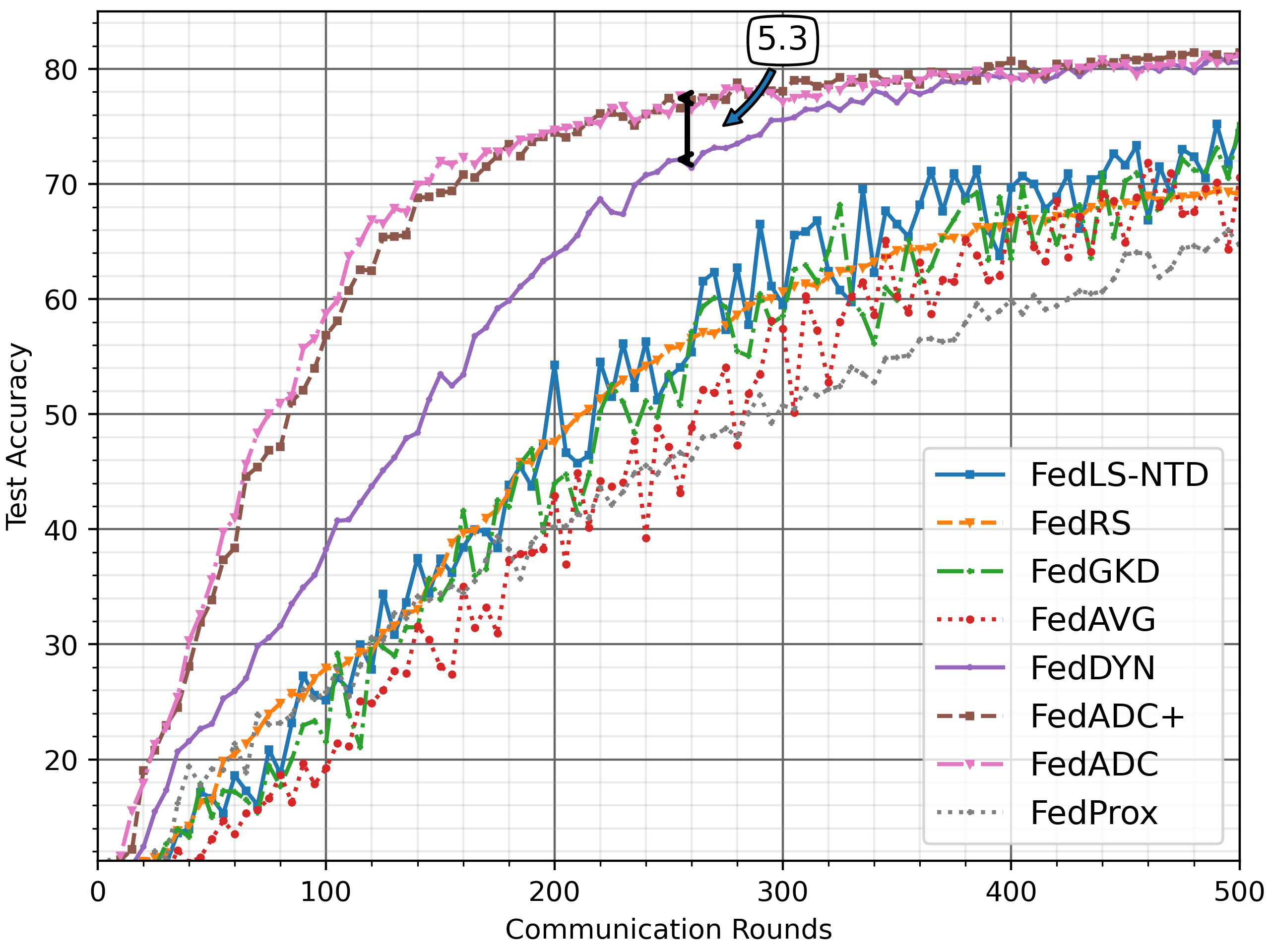}
         \caption{$C=0.2$}
         \label{cif10-02}
     \end{subfigure}
     \hfill
        \caption{Performance comparison of different schemes for CNN architecture at Cifar-10 dataset with $s=2$ sort and partition data distribution.}
        \label{fig:CIFAR-10-drift}
\end{figure}

\begin{figure}[h!]
    \begin{subfigure}{0.49\textwidth}
        \includegraphics[width=\linewidth]{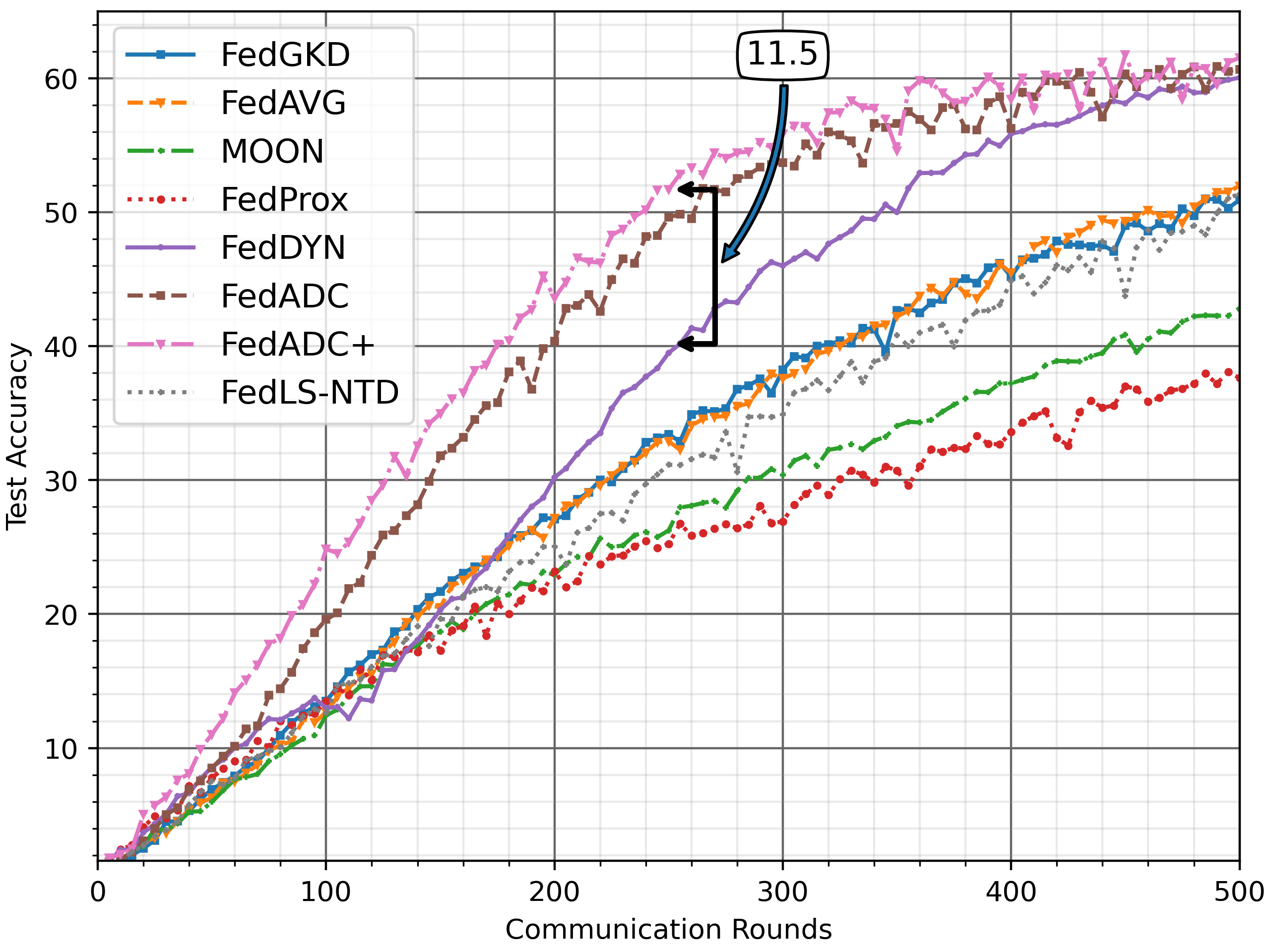}
        \caption{$C=0.2$ and $\alpha=0.1$}
        \label{fig:0102}
    \end{subfigure}\hspace{1em}%
    \begin{subfigure}{0.49\textwidth}
        \includegraphics[width=\linewidth]{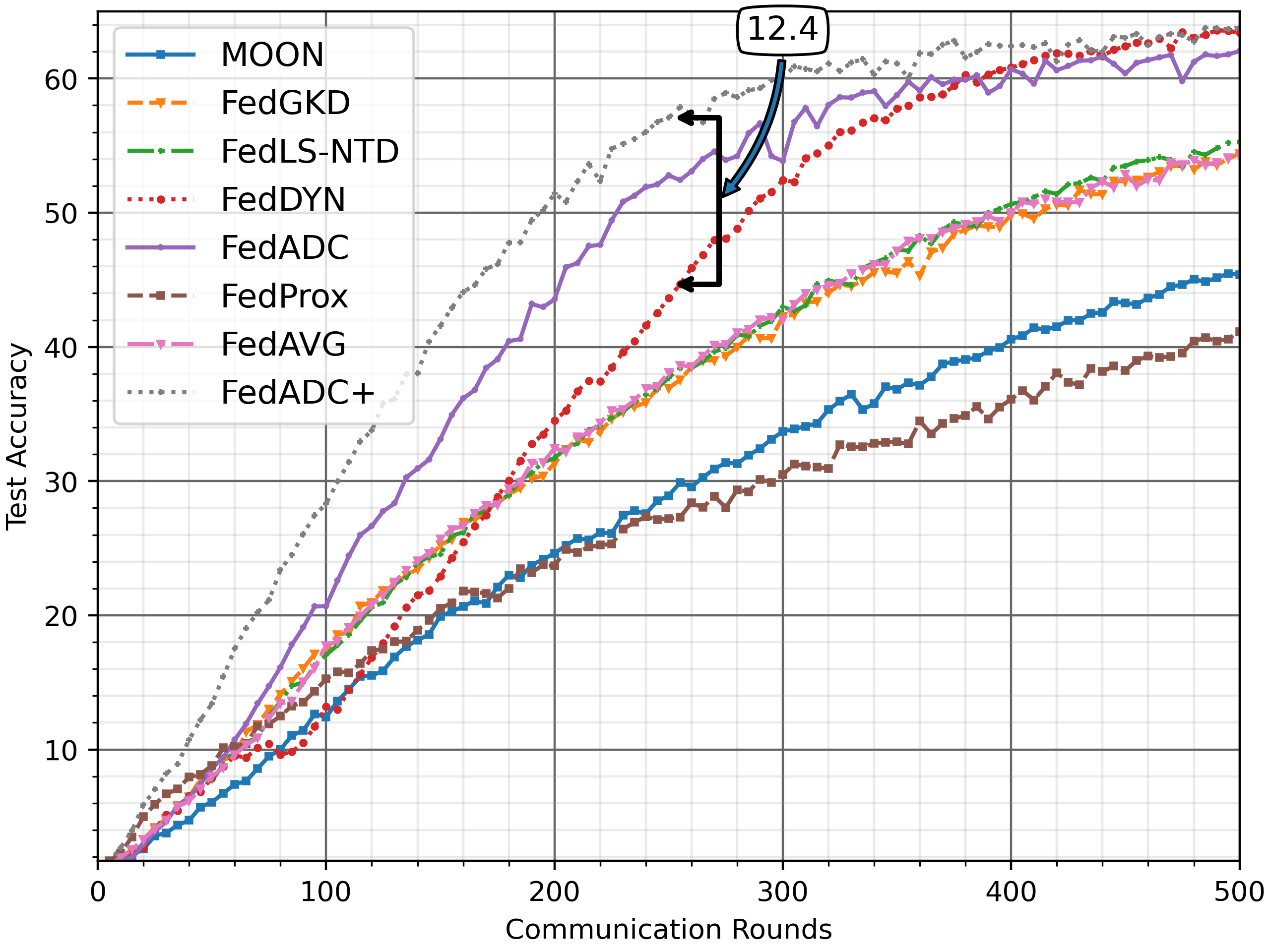}
        \caption{$C=0.2$ and $\alpha=0.5$}
        \label{fig:0502}
    \end{subfigure}
    \begin{subfigure}{0.49\textwidth}
        \includegraphics[width=\linewidth]{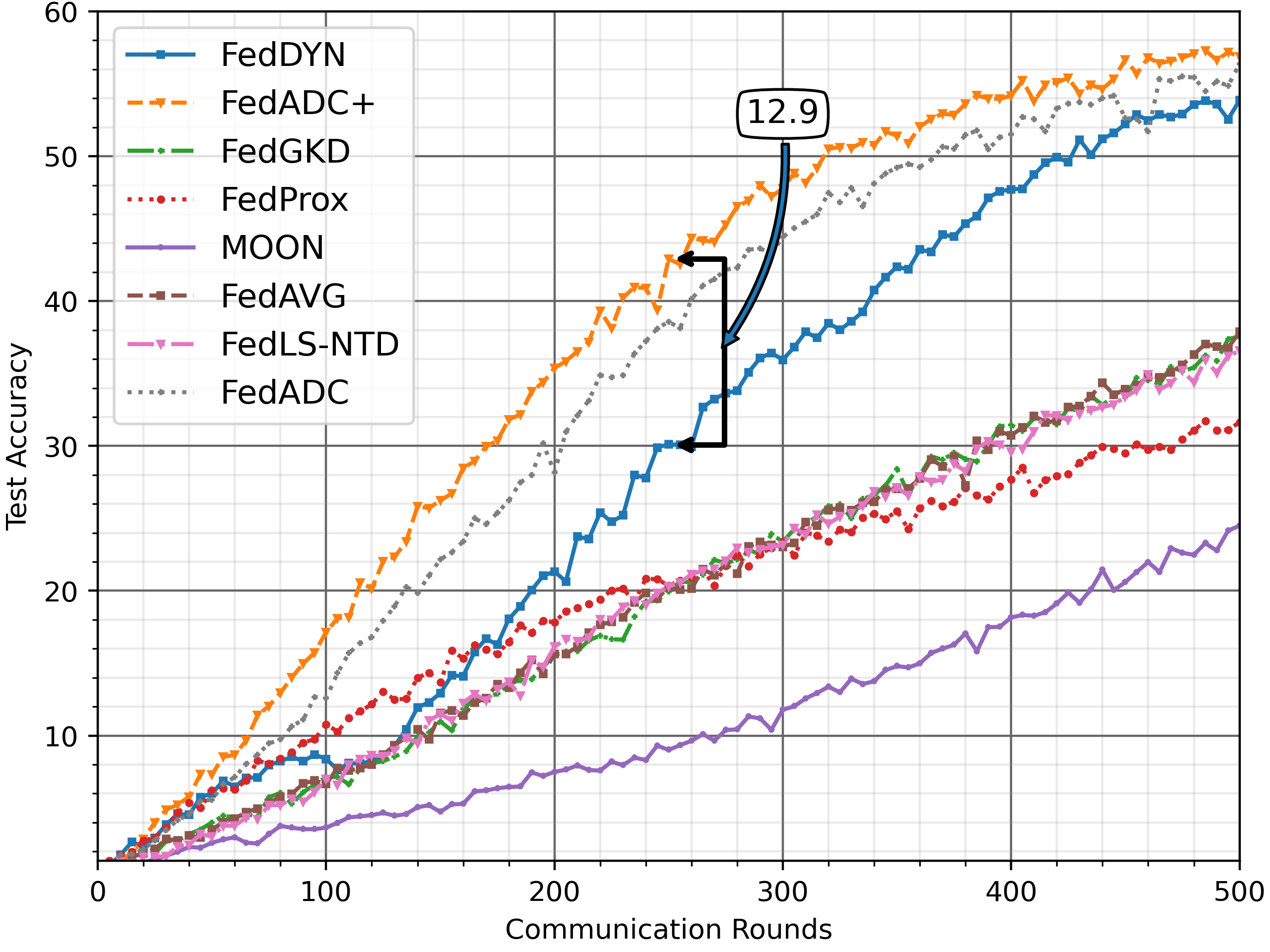}
        \caption{$C=0.1$ and $\alpha=0.1$}
        \label{fig:0101}
    \end{subfigure}\hspace{1em}%
    \begin{subfigure}{0.49\textwidth}
        \includegraphics[width=\linewidth]{0501.png}
        \caption{$C=0.1$ and $\alpha=0.5$}
        \label{fig:0501}
    \end{subfigure}
    \caption{Performance comparisons of different schemes on ResNet-18 DNN architecture at Cifar-100 dataset with  Dirichlet distributions.}
    \label{fig:cifar-100}

\end{figure}

\begin{figure}
     \centering
     \begin{subfigure}[b]{0.32\textwidth}
         \centering
         \includegraphics[width=\textwidth]{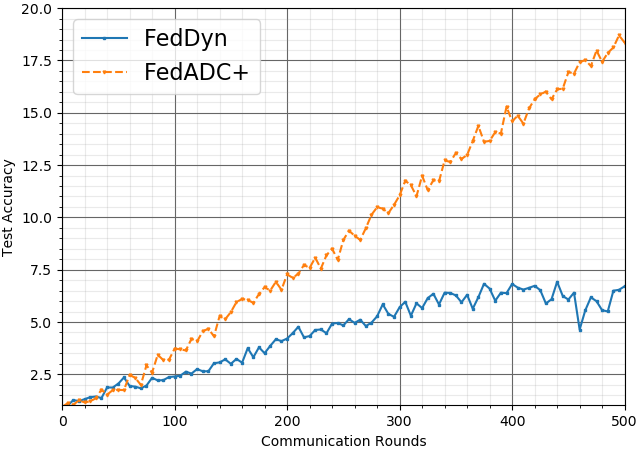}
         \caption{2 local epoch}
         \label{fig:500_2}
     \end{subfigure}
     \hfill
     \begin{subfigure}[b]{0.32\textwidth}
         \centering
         \includegraphics[width=\textwidth]{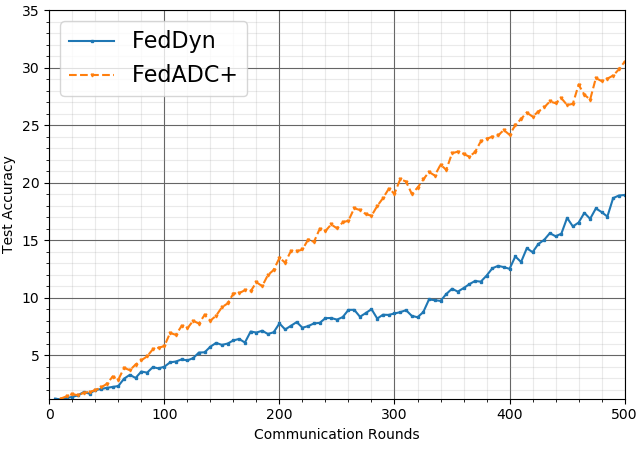}
         \caption{5 local epoch}
         \label{fig:500_5}
     \end{subfigure}
     \hfill
     \begin{subfigure}[b]{0.32\textwidth}
         \centering
         \includegraphics[width=\textwidth]{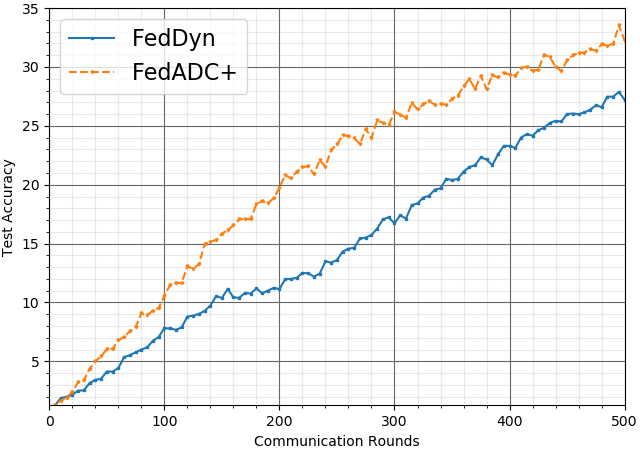}
         \caption{10 local epoch}
         \label{fig:500_10}
     \end{subfigure}
        \caption{Performance comparisons of FedADC+ and FedDyn on ResNet-18 for Cifar-100 dataset with Dirichlet distribution of $\alpha=0.1$. 500 total clients with a $C=0.02$ participation ratio.}
        \label{fig:500cl}
\end{figure}

\begin{figure}[]
\centering
\includegraphics[scale=0.5]{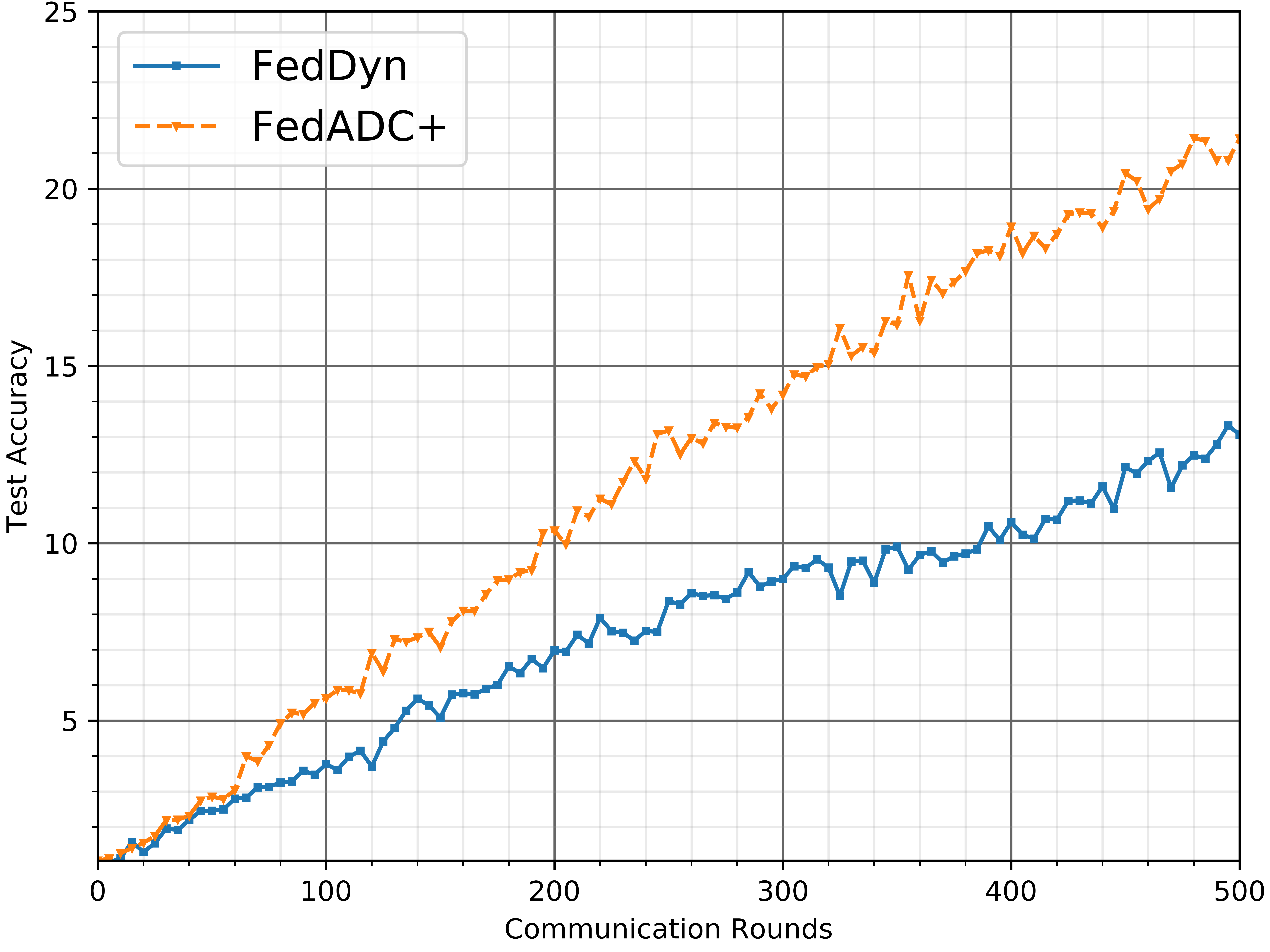}
\caption{Performance comparison of FedADC+ and FedDyn on ResNet-18 for Cifar-100 dataset with Dirichlet distribution of $\alpha=0.1$. 1000 total clients with a $C=0.01$ participation ratio and 10 local epochs of training.}
\label{fig:1000}
\end{figure}

In table \ref{tab:all_results} we compare FedADC and FedADC+ with most prominent FL algorithms in the current literature. In order to demonstrate the capabilities and robustness of our algorithm, we create 6 different scenarios involving total of 100 client with 2 different dataset, 2 different data distribution schemes and finally 2 different client participation ratio. In Table \ref{tab:all_results}, we train all the experiments for $500$ communication rounds,  each of which consists of $2$ local epoch, $H\cong10$. We illustrate the convergence behavior for Table \ref{tab:all_results} at Fig. \ref{fig:CIFAR-10-drift} for Cifar-10 and Fig. \ref{fig:cifar-100} for Cifar-100 respectively with an additional emphasis on the acceleration of FedADC algorithms by showing difference with its nearest competitor when it reaches to half of the total communication rounds. To further illustrate the versatility and robustness of FedADC algorithm, we include 2 more extreme FL scenarios for Cifar-100 image classification problem where we compare FedADC+ with its closest competitor FedDyn \cite{FedDyn}, on dirichlet distribution of $\alpha=0.1$. For this scenarios, we increase the total number of client to 500 and 1000 while fixing the total participating client number to $10$ for the each round. In Fig. \ref{fig:500cl}, we illustrate convergence of the both algorithms with an emphasis on their local epoch numbers. We show that FedADC+ can provide much better convergence and a higher final test accuracy regardless of number of the local epoch while FedDyn struggles to maintain a normal convergence at Fig. \ref{fig:500_2} and Fig. \ref{fig:500_5} which corresponds to the 2 and 5 local epochs respectively. In Fig. \ref{fig:1000}, we increase the number of clients to 1000 to further demonstrate that regardless of the total client number, FedADC+ sustained the faster convergence while also surpassing the FedDyn in the final test accuracy.

\begin{table}[]
\centering
\begin{tabular}{lllllll}
\hline
\multicolumn{1}{c}{\textbf{Algorithm}} & \textbf{\begin{tabular}[c]{@{}l@{}}CIFAR-10\\ $s$=2 C=0.2\end{tabular}} & \textbf{\begin{tabular}[c]{@{}l@{}}CIFAR-10\\ $s$=2 C=0.1\end{tabular}} & \textbf{\begin{tabular}[c]{@{}l@{}}CIFAR-100\\ $\alpha$=0.5 C=0.2\end{tabular}} & \textbf{\begin{tabular}[c]{@{}l@{}}CIFAR-100\\ $\alpha$=0.1 C=0.2\end{tabular}} & \textbf{\begin{tabular}[c]{@{}l@{}}CIFAR-100\\ $\alpha$=0.5 C=0.1\end{tabular}} & \textbf{\begin{tabular}[c]{@{}l@{}}CIFAR-100\\ $\alpha$=0.1 C=0.1\end{tabular}} \\ \hline
FedAVG\cite{FedAVG} & 70.55  $\pm 2.5$ & 65.16 $\pm 4.1$ & 54.22 $\pm 1.08$ & 43.88 $\pm 1.4$ & 51.2 $\pm 1.55$ & 35.21 $\pm 1.51$ \\ \hline
MOON \cite{Moon} & 51.53 $\pm 2.05$ & 44.83 $\pm 1.83$ & 45.38 $\pm 1.21$ & 29.32 $\pm 0.56$ & 42.83 $\pm 0.4$ & 24.5 $\pm 1.31$ \\ \hline
FedGKD \cite{dist2}& 75.28 $\pm 1.28$ & 66 $\pm 1.28$ & 54.41 $\pm 1.15$ & 44.32 $\pm 2.3$ & 50.93 $\pm 0.57$ & 37.54 $\pm 1.95$ \\ \hline
FedNTD \cite{dist1} & 74.35 $\pm 0.22$ & 66.51 $\pm 5.35$ & 55.29 $\pm 1.51$ & 44.71 $\pm 1.1$ & 51.38 $\pm 0.74$ & 36.57 $\pm 1.75$ \\ \hline
FedDyn \cite{FedDyn} & 80.58 $\pm 1.2$ & 77.34 $\pm 1.13$ & {\ul 63.4 $\pm 0.67$} & {\ul 58.49 $\pm 0.32$} & 60.5 $\pm 0.68$ & 53.86 $\pm 0.19$ \\ \hline
FedProx \cite{fedprox} & 64.67 $\pm 1.63$  & 63.13 $\pm 4.48$ & 41.15 $\pm 1.06$ & 33.85 $\pm 0.79$ & 37.62 $\pm 0.82$ & 31.63 $\pm 0.33$\\ \hline
\textbf{FedADC (ours)} & {\ul 81.13 $\pm 0.82$ } & {\ul 78.62 $\pm 0.94$ } & 62.04 $\pm 0.48$ & 57.91  $\pm 0.55$& {\ul 60.66 $\pm 0.24$} & {\ul 56.37 $\pm 1.06$} \\ \hline
\textbf{FedADC+ (ours)} & \textbf{81.42 $\pm $0.35} & \textbf{78.77 $\pm$ 0.7} & \textbf{63.8 $\pm$ 0.29 } & \textbf{59.75 $\pm $ 0.9 } & \textbf{61.55 $\pm$ 0.85 } & \textbf{56.9 $\pm$ 0.67} \\ \hline
FedRS \cite{FedRS}& 69.1 $\pm 1.9$ & 66.55 $\pm 1.77$ & - & - & - & - \\ \hline
\end{tabular}
\caption{Top-1 Test Accuracy overview for given different data and participation settings on Cifar Datasets (100 clients, 500 communication rounds with 2 local epoch training. \textbf{Bold} for best result, {\ul underlined} for runner-up results)}
\label{tab:all_results}
\end{table}

\subsection{Personalization with classifier calibration}
Although, in the scope of this work we primarily focus on the generalization capability of FL, another important aspect often addressed in the literature is the personalization of the FL framework \cite{personalized1,personalized2,personalized3,personalized4,personalized5,personalized6,personalized7,personalized8,personalized9,personalized10,personalized11}. Although, there are several different strategies to personalize the the learned model, we can identify two prominent strategies. The first strategy is to update global and local model 
in parallel during the FL \cite{personalized4,personalized5} to mitigate the erasure of the local information. The major drawback of such strategies is that both the computational and memory requirements twice of the naive FedAvg. The second, alternative strategy is to personalize the model such that the  layers of the network is divided in to two groups and the certain layers are only trained/fine-tuned locally \cite{personalized8,personalized9,personalized10}. The common approach is to divide the network  into parts called {\em body} and {\em head} which correspond to lower and higher layers in the architectures. Then, in a  broad sense, the body is trained globally to learn to extract meaningful features and train the head locally to achieve  personalized decision according to local dataset. One particular implementation of such strategy can be referred as {\em classifier calibration}\cite{calib1,calib2,calib3}, where the head contains the last layer and personalization is achieved by tuning the corresponding classifier layer using the local dataset. Such strategy is highly efficient in terms of both communication and computation, and yet quite effective as we empirically demonstrate later.

\begin{figure}[]
\centering
\includegraphics[scale=0.5]{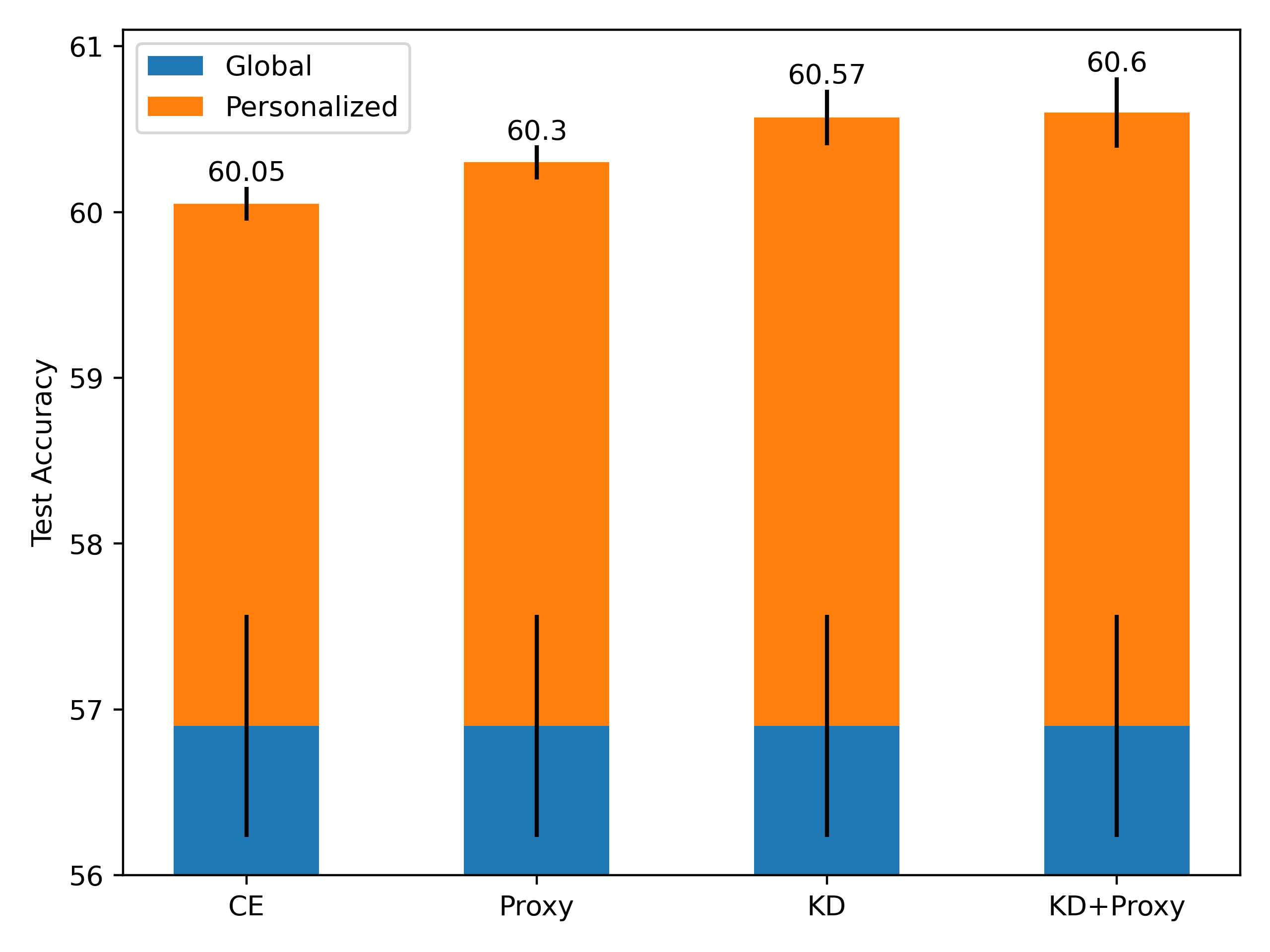}
\caption{The mean average accuracy of FedADC+ with personalization on CIFAR-100 image classification task for $C=0.1$ and $\alpha=0.1$.}
\label{fig:personal}
\end{figure}

For simulation, we chose image classification task for CIFAR-100 dataset with Dir($\alpha=0.1$) dirichlet distribution for both training and test dataset. All local clients make only local computations with their own train set and reported their test accuracy respective to their share of the local test set. For the training, each client starts with global model then continue to training locally for 2 local epochs with only computing the gradients for the classifier layer. We also consider two different regularization methods to further improve the test accuracy. We use proxy method introduced at \cite{fedprox}, where it adds the regularization of proximal term for local training and the self-knowledge distillation at section \ref{conf-KD} which we also consider it for the federated learning. We illustrate the results at Fig \ref{fig:personal} where we are able surpass the cumulative test accuracy of the global model by $\%3.3- \% 4.1$. 

Finally, we emphasize here that, unlike many of the existing personalized FL schemes proposed, due to its structure the proposed approach is robust to the changes in the local datasets, that is the classifier calibration can be easily repeated to adapt the new local data statistics.

\subsection{Further discussion}
For the completeness of the related works, we note that use of local and global momentum terms has been also discussed in \cite{lgm1,lgm2,lgm3,lgm4}, however, the proposed schemes either increase the communication overhead or the requires extra computation, in some schemes both, compared to the FedAvg framework which prevent fair comparisons.

\indent One of the key advantages of the proposed FedADC scheme is that it can be employed in large scale scenarios with low client participation ratio which is desired for practical scenarios. The success of the FedADC scheme is mainly due to the regularizing affect of the locally embedded global momentum term. On the other hand, embedding the global momentum leads to a biased updates particularly when only small number of clients are scheduled for the model update. To be more precise, when the number of the scheduled clients are small, we expect the distribution of the union of the local dataset of the scheduled clients also to be non-iid, thus the momentum update is biased. \\
Hence, to make the union of the local dataset of the scheduled clients more uniform without increasing the participation ratio, clients can be clustered based on similarity of their local dataset, and, instead of random sampling, clients can be scheduled in a such way union of the local dataset have more uniform distribution which may help to mitigate any bias on the momentum term. We note that clustering framework have been already discussed in \cite{clusterFL1,clusterFL2} to improve the performance of the FL framework. To illustrate the possible improvements with client clustering to perform data-aware client selection, we revisit the simulation setup in section \ref{sec:drift_setup} specifically, image classification task for Cifar-10 dataset with $s=2$ and $C=0.1$. For this simulation, in each round, parameter server chooses random subset of clients with additional constrain where accumulated training data from the clients must include a sample from the all the classes of the dataset. Asserting a such clustering scheme further improves the final test accuracy by $\%2.1$.\\
\indent Finally, the accuracy of the global model can be further improved by ensembling models over the time following the proposed strategies in \cite{swa} and \cite{lookahead}. The temporal ensemble of the global models can be also used for regularization of the local updates at client \cite{dist2} or to improve the ensembling at PS \cite{ensemble2}.

\section{Conclusion}
In this paper, we introduced the FedADC scheme to make the FL framework more robust to data heterogeneity among the users. The proposed strategy embeds the momentum update step used at the server side into the local model update procedure to control the local drift and prevent divergence. Through experiments on a  CNN and ResNet architecture for image classification on the CIFAR-10 and CIFAR-100 datasets, we show that the proposed FedADC and FedADC+ approach accelerates the training while also  preventing  local drifts, and as a result, outperforms the benchmark FedAvg and SlOWMO strategies as well as SOTA strategies in terms of the test accuracy and convergence performance.

\bibliographystyle{IEEEtran}
\bibliography{IEEEabrv,ref.bib}

\begin{thebibliography}{10}
\providecommand{\url}[1]{#1}
\csname url@samestyle\endcsname
\providecommand{\newblock}{\relax}
\providecommand{\bibinfo}[2]{#2}
\providecommand{\BIBentrySTDinterwordspacing}{\spaceskip=0pt\relax}
\providecommand{\BIBentryALTinterwordstretchfactor}{4}
\providecommand{\BIBentryALTinterwordspacing}{\spaceskip=\fontdimen2\font plus
\BIBentryALTinterwordstretchfactor\fontdimen3\font minus
  \fontdimen4\font\relax}
\providecommand{\BIBforeignlanguage}[2]{{%
\expandafter\ifx\csname l@#1\endcsname\relax
\typeout{** WARNING: IEEEtran.bst: No hyphenation pattern has been}%
\typeout{** loaded for the language `#1'. Using the pattern for}%
\typeout{** the default language instead.}%
\else
\language=\csname l@#1\endcsname
\fi
#2}}
\providecommand{\BIBdecl}{\relax}
\BIBdecl

\bibitem{FL1}
B.~McMahan, E.~Moore, D.~Ramage, S.~Hampson, and B.~A. y~Arcas,
  ``{Communication-Efficient Learning of Deep Networks from Decentralized
  Data},'' in \emph{Proceedings of the 20th International Conference on
  Artificial Intelligence and Statistics}, ser. Proceedings of Machine Learning
  Research, vol.~54.\hskip 1em plus 0.5em minus 0.4em\relax Fort Lauderdale,
  FL, USA: PMLR, Apr 2017, pp. 1273--1282.

\bibitem{FL.keyboard1}
A.~Hard, K.~Rao, R.~Mathews, F.~Beaufays, S.~Augenstein, H.~Eichner, C.~Kiddon,
  and D.~Ramage, ``Federated learning for mobile keyboard prediction,''
  \emph{CoRR}, vol. abs/1811.03604, 2018.

\bibitem{FL.keyboard2}
S.~Ramaswamy, R.~Mathews, K.~Rao, and F.~Beaufays, ``Federated learning for
  emoji prediction in a mobile keyboard,'' \emph{CoRR}, vol. abs/1906.04329,
  2019.

\bibitem{FL.health1}
W.~Li, F.~Milletar{\`i}, D.~Xu, N.~Rieke, J.~Hancox, W.~Zhu, M.~Baust,
  Y.~Cheng, S.~Ourselin, M.~J. Cardoso, and A.~Feng, ``Privacy-preserving
  federated brain tumour segmentation,'' in \emph{Machine Learning in Medical
  Imaging}.\hskip 1em plus 0.5em minus 0.4em\relax Cham: Springer International
  Publishing, 2019, pp. 133--141.

\bibitem{FL.health2}
N.~Rieke, J.~Hancox, W.~Li, F.~Milletari, H.~Roth, S.~Albarqouni, S.~Bakas,
  M.~N. Galtier, B.~Landman, K.~Maier-Hein, S.~Ourselin, M.~Sheller, R.~M.
  Summers, A.~Trask, D.~Xu, M.~Baust, and M.~J. Cardoso, ``The future of
  digital health with federated learning,'' \emph{CoRR}, vol. abs/2003.08119,
  2020.

\bibitem{FL.health3}
\BIBentryALTinterwordspacing
M.~Malekzadeh, B.~Hasircioglu, N.~Mital, K.~Katarya, M.~E. Ozfatura, and
  D.~G{\"{u}}nd{\"{u}}z, ``Dopamine: Differentially private federated learning
  on medical data,'' \emph{CoRR}, vol. abs/2101.11693, 2021. [Online].
  Available: \url{https://arxiv.org/abs/2101.11693}
\BIBentrySTDinterwordspacing

\bibitem{FL.noniid1}
T.~H. Hsu, H.~Qi, and M.~Brown, ``Measuring the effects of non-identical data
  distribution for federated visual classification,'' \emph{CoRR}, vol.
  abs/1909.06335, 2019.

\bibitem{FL.noniid2}
T.-M.~H. Hsu, H.~Qi, and M.~Brown, ``Federated visual classification with
  real-world data distribution,'' \emph{CoRR}, vol. abs/2003.08082, 2020.

\bibitem{FL.noniid3}
K.~Hsieh, A.~Phanishayee, O.~Mutlu, and P.~Gibbons, ``The non-{IID} data
  quagmire of decentralized machine learning,'' in \emph{Proceedings of the
  37th International Conference on Machine Learning}, ser. Proceedings of
  Machine Learning Research, vol. 119.\hskip 1em plus 0.5em minus 0.4em\relax
  Virtual: PMLR, 13--18 Jul 2020, pp. 4387--4398.

\bibitem{FL.noniid4}
Y.~Zhao, M.~Li, L.~Lai, N.~Suda, D.~Civin, and V.~Chandra, ``Federated learning
  with non-iid data,'' \emph{CoRR}, vol. abs/1806.00582, 2018.

\bibitem{scaffold}
S.~P. Karimireddy, S.~Kale, M.~Mohri, S.~Reddi, S.~Stich, and A.~T. Suresh,
  ``{SCAFFOLD}: Stochastic controlled averaging for federated learning,'' in
  \emph{Proceedings of the 37th International Conference on Machine Learning},
  ser. Proceedings of Machine Learning Research, H.~D. III and A.~Singh, Eds.,
  vol. 119.\hskip 1em plus 0.5em minus 0.4em\relax Virtual: PMLR, 13--18 Jul
  2020, pp. 5132--5143.

\bibitem{fedprox}
T.~Li, A.~K. Sahu, M.~Zaheer, M.~Sanjabi, A.~Talwalkar, and V.~Smith,
  ``Federated optimization in heterogeneous networks,'' in \emph{Proceedings of
  Machine Learning and Systems}, I.~Dhillon, D.~Papailiopoulos, and V.~Sze,
  Eds., vol.~2, 2020, pp. 429--450.

\bibitem{SGD.svr}
\BIBentryALTinterwordspacing
S.~J. Reddi, A.~Hefny, S.~Sra, B.~Poczos, and A.~Smola, ``Stochastic variance
  reduction for nonconvex optimization,'' ser. Proceedings of Machine Learning
  Research, M.~F. Balcan and K.~Q. Weinberger, Eds., vol.~48.\hskip 1em plus
  0.5em minus 0.4em\relax New York, New York, USA: PMLR, 20--22 Jun 2016, pp.
  314--323. [Online]. Available:
  \url{http://proceedings.mlr.press/v48/reddi16.html}
\BIBentrySTDinterwordspacing

\bibitem{SGD.nesterov}
Y.~Nesterov, ``A method for solving the convex programming problem with
  convergence rate $o(1/k^2)$,'' \emph{Proceedings of the USSR Academy of
  Sciences}, vol. 269, pp. 543--547, 1983.

\bibitem{SGD.polyak}
B.~Polyak, ``Some methods of speeding up the convergence of iteration
  methods,'' \emph{USSR Computational Mathematics and Mathematical Physics},
  vol.~4, no.~5, pp. 1 -- 17, 1964.

\bibitem{SGD.opt2}
I.~Sutskever, J.~Martens, G.~Dahl, and G.~Hinton, ``On the importance of
  initialization and momentum in deep learning,'' ser. Proceedings of Machine
  Learning Research, vol.~28, no.~3.\hskip 1em plus 0.5em minus 0.4em\relax
  Atlanta, Georgia, USA: PMLR, 17--19 Jun 2013, pp. 1139--1147.

\bibitem{FL.acc1}
J.~Wang, V.~Tantia, N.~Ballas, and M.~Rabbat, ``{SlOWMO}: Improving
  communication-efficient distributed {SGD} with slow momentum,'' in
  \emph{International Conference on Learning Representations}, 2020.

\bibitem{FL.acc2}
S.~Reddi, Z.~Charles, M.~Zaheer, Z.~Garrett, K.~Rush, J.~Konečný, S.~Kumar,
  and H.~B. McMahan, ``Adaptive federated optimization,'' \emph{CoRR}, vol.
  abs/2003.00295, 2020.

\bibitem{FL.acc3}
K.~{Chen}, H.~{Ding}, and Q.~{Huo}, ``Parallelizing adam optimizer with
  blockwise model-update filtering,'' in \emph{IEEE International Conference on
  Acoustics, Speech and Signal Processing (ICASSP)}, 2020, pp. 3027--3031.

\bibitem{lookahead}
M.~Zhang, J.~Lucas, J.~Ba, and G.~E. Hinton, ``Lookahead optimizer: k steps
  forward, 1 step back,'' in \emph{Advances in Neural Information Processing
  Systems 32}.\hskip 1em plus 0.5em minus 0.4em\relax Curran Associates, Inc.,
  2019, pp. 9597--9608.

\bibitem{SGD.opt1}
J.-K. Wang, C.-H. Lin, and J.~Abernethy, ``Escaping saddle points faster with
  stochastic momentum,'' in \emph{International Conference on Learning
  Representations}, 2020.

\bibitem{selfkd1}
X.~Wang, Y.~Hua, E.~Kodirov, D.~A. Clifton, and N.~M. Robertson, ``Proselflc:
  Progressive self label correction for training robust deep neural networks,''
  in \emph{Proceedings of the IEEE/CVF Conference on Computer Vision and
  Pattern Recognition (CVPR)}, June 2021, pp. 752--761.

\bibitem{selfkd2}
S.~Yun, J.~Park, K.~Lee, and J.~Shin, ``Regularizing class-wise predictions via
  self-knowledge distillation,'' in \emph{Proceedings of the IEEE/CVF
  Conference on Computer Vision and Pattern Recognition (CVPR)}, June 2020.

\bibitem{selfkd3}
L.~Yuan, F.~E. Tay, G.~Li, T.~Wang, and J.~Feng, ``Revisiting knowledge
  distillation via label smoothing regularization,'' in \emph{Proceedings of
  the IEEE/CVF Conference on Computer Vision and Pattern Recognition (CVPR)},
  June 2020.

\bibitem{selfkd4}
K.~Kim, B.~Ji, D.~Yoon, and S.~Hwang, ``Self-knowledge distillation with
  progressive refinement of targets,'' in \emph{Proceedings of the IEEE/CVF
  International Conference on Computer Vision (ICCV)}, October 2021, pp.
  6567--6576.

\bibitem{dist1}
\BIBentryALTinterwordspacing
G.~Lee, Y.~Shin, M.~Jeong, and S.~Yun, ``Preservation of the global knowledge
  by not-true self knowledge distillation in federated learning,'' \emph{CoRR},
  vol. abs/2106.03097, 2021. [Online]. Available:
  \url{https://arxiv.org/abs/2106.03097}
\BIBentrySTDinterwordspacing

\bibitem{dist2}
\BIBentryALTinterwordspacing
W.~Pan and L.~Sun, ``Global knowledge distillation in federated learning,''
  \emph{CoRR}, vol. abs/2107.00051, 2021. [Online]. Available:
  \url{https://arxiv.org/abs/2107.00051}
\BIBentrySTDinterwordspacing

\bibitem{distadv}
\BIBentryALTinterwordspacing
T.~Chen, Z.~Zhang, S.~Liu, S.~Chang, and Z.~Wang, ``Robust overfitting may be
  mitigated by properly learned smoothening,'' in \emph{International
  Conference on Learning Representations}, 2021. [Online]. Available:
  \url{https://openreview.net/forum?id=qZzy5urZw9}
\BIBentrySTDinterwordspacing

\bibitem{ensemble1}
T.~Lin, L.~Kong, S.~U. Stich, and M.~Jaggi, ``Ensemble distillation for robust
  model fusion in federated learning,'' in \emph{Proceedings of the 34th
  International Conference on Neural Information Processing Systems}, ser.
  NIPS'20.\hskip 1em plus 0.5em minus 0.4em\relax Red Hook, NY, USA: Curran
  Associates Inc., 2020.

\bibitem{ensemble3}
\BIBentryALTinterwordspacing
Z.~Zhu, J.~Hong, and J.~Zhou, ``Data-free knowledge distillation for
  heterogeneous federated learning,'' \emph{CoRR}, vol. abs/2105.10056, 2021.
  [Online]. Available: \url{https://arxiv.org/abs/2105.10056}
\BIBentrySTDinterwordspacing

\bibitem{FedRS}
\BIBentryALTinterwordspacing
X.-C. Li and D.-C. Zhan, \emph{FedRS: Federated Learning with Restricted
  Softmax for Label Distribution Non-IID Data}.\hskip 1em plus 0.5em minus
  0.4em\relax New York, NY, USA: Association for Computing Machinery, 2021, p.
  995–1005. [Online]. Available:
  \url{https://doi.org/10.1145/3447548.3467254}
\BIBentrySTDinterwordspacing

\bibitem{Moon}
Q.~Li, B.~He, and D.~Song, ``Model-contrastive federated learning,'' in
  \emph{Proceedings of the IEEE/CVF Conference on Computer Vision and Pattern
  Recognition (CVPR)}, June 2021, pp. 10\,713--10\,722.

\bibitem{FedDyn}
D.~A.~E. Acar, Y.~Zhao, R.~M. Navarro, M.~Mattina, P.~N. Whatmough, and
  V.~Saligrama, ``Federated learning based on dynamic regularization,'' 2021.

\bibitem{cifar10}
\BIBentryALTinterwordspacing
A.~Krizhevsky, V.~Nair, and G.~Hinton, ``Cifar-10 (canadian institute for
  advanced research).'' [Online]. Available:
  \url{http://www.cs.toronto.edu/~kriz/cifar.html}
\BIBentrySTDinterwordspacing

\bibitem{ResNet}
\BIBentryALTinterwordspacing
K.~He, X.~Zhang, S.~Ren, and J.~Sun, ``Deep residual learning for image
  recognition,'' \emph{CoRR}, vol. abs/1512.03385, 2015. [Online]. Available:
  \url{http://arxiv.org/abs/1512.03385}
\BIBentrySTDinterwordspacing

\bibitem{GroupNorm}
\BIBentryALTinterwordspacing
Y.~Wu and K.~He, ``Group normalization,'' \emph{CoRR}, vol. abs/1803.08494,
  2018. [Online]. Available: \url{http://arxiv.org/abs/1803.08494}
\BIBentrySTDinterwordspacing

\bibitem{FedAVG}
B.~McMahan, E.~Moore, D.~Ramage, S.~Hampson, and B.~A. y~Arcas,
  ``Communication-efficient learning of deep networks from decentralized
  data,'' in \emph{Artificial intelligence and statistics}.\hskip 1em plus
  0.5em minus 0.4em\relax PMLR, 2017, pp. 1273--1282.

\bibitem{personalized1}
A.~Fallah, A.~Mokhtari, and A.~Ozdaglar, ``Personalized federated learning: A
  meta-learning approach,'' 2020.

\bibitem{personalized2}
\BIBentryALTinterwordspacing
M.~Zhang, K.~Sapra, S.~Fidler, S.~Yeung, and J.~M. Alvarez, ``Personalized
  federated learning with first order model optimization,'' in
  \emph{International Conference on Learning Representations}, 2021. [Online].
  Available: \url{https://openreview.net/forum?id=ehJqJQk9cw}
\BIBentrySTDinterwordspacing

\bibitem{personalized3}
\BIBentryALTinterwordspacing
C.~T.~Dinh, N.~Tran, and J.~Nguyen, ``Personalized federated learning with
  moreau envelopes,'' in \emph{Advances in Neural Information Processing
  Systems}, H.~Larochelle, M.~Ranzato, R.~Hadsell, M.~F. Balcan, and H.~Lin,
  Eds., vol.~33.\hskip 1em plus 0.5em minus 0.4em\relax Curran Associates,
  Inc., 2020, pp. 21\,394--21\,405. [Online]. Available:
  \url{https://proceedings.neurips.cc/paper/2020/file/f4f1f13c8289ac1b1ee0ff176b56fc60-Paper.pdf}
\BIBentrySTDinterwordspacing

\bibitem{personalized4}
\BIBentryALTinterwordspacing
T.~Li, S.~Hu, A.~Beirami, and V.~Smith, ``Ditto: Fair and robust federated
  learning through personalization,'' in \emph{Proceedings of the 38th
  International Conference on Machine Learning}, ser. Proceedings of Machine
  Learning Research, M.~Meila and T.~Zhang, Eds., vol. 139.\hskip 1em plus
  0.5em minus 0.4em\relax PMLR, 18--24 Jul 2021, pp. 6357--6368. [Online].
  Available: \url{https://proceedings.mlr.press/v139/li21h.html}
\BIBentrySTDinterwordspacing

\bibitem{personalized5}
Y.~Deng, M.~M. Kamani, and M.~Mahdavi, ``Adaptive personalized federated
  learning,'' 2020.

\bibitem{personalized6}
\BIBentryALTinterwordspacing
Y.~Jiang, J.~Kone{\v{c}}n{\'y}, K.~Rush, and S.~Kannan, ``Improving federated
  learning personalization via model agnostic meta learning,'' 2020. [Online].
  Available: \url{https://openreview.net/forum?id=BkeaEyBYDB}
\BIBentrySTDinterwordspacing

\bibitem{personalized7}
\BIBentryALTinterwordspacing
Y.~Mansour, M.~Mohri, J.~Ro, and A.~T. Suresh, ``Three approaches for
  personalization with applications to federated learning,'' \emph{CoRR}, vol.
  abs/2002.10619, 2020. [Online]. Available:
  \url{https://arxiv.org/abs/2002.10619}
\BIBentrySTDinterwordspacing

\bibitem{personalized8}
\BIBentryALTinterwordspacing
M.~G. Arivazhagan, V.~Aggarwal, A.~K. Singh, and S.~Choudhary, ``Federated
  learning with personalization layers,'' \emph{CoRR}, vol. abs/1912.00818,
  2019. [Online]. Available: \url{http://arxiv.org/abs/1912.00818}
\BIBentrySTDinterwordspacing

\bibitem{personalized9}
L.~Collins, H.~Hassani, A.~Mokhtari, and S.~Shakkottai, ``Exploiting shared
  representations for personalized federated learning,'' 2021.

\bibitem{personalized10}
K.~Singhal, H.~Sidahmed, Z.~Garrett, S.~Wu, K.~Rush, and S.~Prakash,
  ``Federated reconstruction: Partially local federated learning,'' 2021.

\bibitem{personalized11}
P.~P. Liang, T.~Liu, L.~Ziyin, N.~B. Allen, R.~P. Auerbach, D.~Brent,
  R.~Salakhutdinov, and L.-P. Morency, ``Think locally, act globally: Federated
  learning with local and global representations,'' 2020.

\bibitem{calib1}
J.~Oh, S.~Kim, and S.-Y. Yun, ``Fedbabu: Towards enhanced representation for
  federated image classification,'' 2021.

\bibitem{calib2}
M.~Luo, F.~Chen, D.~Hu, Y.~Zhang, J.~Liang, and J.~Feng, ``No fear of
  heterogeneity: Classifier calibration for federated learning with non-iid
  data,'' 2021.

\bibitem{calib3}
T.~Yu, E.~Bagdasaryan, and V.~Shmatikov, ``Salvaging federated learning by
  local adaptation,'' 2021.

\bibitem{lgm1}
P.~Khanduri, P.~Sharma, H.~Yang, M.~Hong, J.~Liu, K.~Rajawat, and P.~K.
  Varshney, ``Stem: A stochastic two-sided momentum algorithm achieving
  near-optimal sample and communication complexities for federated learning,''
  2021.

\bibitem{lgm2}
R.~Das, A.~Acharya, A.~Hashemi, S.~Sanghavi, I.~S. Dhillon, and U.~Topcu,
  ``Faster non-convex federated learning via global and local momentum,'' 2021.

\bibitem{lgm3}
W.~Liu, L.~Chen, Y.~Chen, and W.~Zhang, ``Accelerating federated learning via
  momentum gradient descent,'' \emph{IEEE Transactions on Parallel and
  Distributed Systems}, vol.~31, no.~8, pp. 1754--1766, 2020.

\bibitem{lgm4}
\BIBentryALTinterwordspacing
H.~Yuan and T.~Ma, ``Federated accelerated stochastic gradient descent,'' in
  \emph{Advances in Neural Information Processing Systems}, H.~Larochelle,
  M.~Ranzato, R.~Hadsell, M.~F. Balcan, and H.~Lin, Eds., vol.~33.\hskip 1em
  plus 0.5em minus 0.4em\relax Curran Associates, Inc., 2020, pp. 5332--5344.
  [Online]. Available:
  \url{https://proceedings.neurips.cc/paper/2020/file/39d0a8908fbe6c18039ea8227f827023-Paper.pdf}
\BIBentrySTDinterwordspacing

\bibitem{clusterFL1}
A.~Ghosh, J.~Chung, D.~Yin, and K.~Ramchandran, ``An efficient framework for
  clustered federated learning,'' 2021.

\bibitem{clusterFL2}
F.~Sattler, K.~M{\"{u}}ller, and W.~Samek, ``Clustered federated learning:
  Model-agnostic distributed multi-task optimization under privacy
  constraints,'' \emph{CoRR}, vol. abs/1910.01991, 2019.

\bibitem{swa}
\BIBentryALTinterwordspacing
P.~Izmailov, D.~Podoprikhin, T.~Garipov, D.~P. Vetrov, and A.~G. Wilson,
  ``Averaging weights leads to wider optima and better generalization,''
  \emph{CoRR}, vol. abs/1803.05407, 2018. [Online]. Available:
  \url{http://arxiv.org/abs/1803.05407}
\BIBentrySTDinterwordspacing

\bibitem{ensemble2}
\BIBentryALTinterwordspacing
H.-Y. Chen and W.-L. Chao, ``Fed{\{}be{\}}: Making bayesian model ensemble
  applicable to federated learning,'' in \emph{International Conference on
  Learning Representations}, 2021. [Online]. Available:
  \url{https://openreview.net/forum?id=dgtpE6gKjHn}
\BIBentrySTDinterwordspacing

\end{thebibliography}

\ifCLASSOPTIONcaptionsoff
  \newpage
\fi





\end{document}